\documentclass{article}

\PassOptionsToPackage{numbers,sort&compress}{natbib}
\usepackage[main,final]{neurips_2026}
\usepackage{enumitem}
\usepackage{caption}
\usepackage{float}
\usepackage{placeins}
\usepackage{amsmath}
\usepackage{booktabs}
\usepackage{multirow}
\usepackage{algorithm}
\usepackage{algorithmic}
\usepackage[table]{xcolor}
\usepackage{tikz}
\definecolor{takeawaybg}{RGB}{235,243,250}
\definecolor{takeawayline}{RGB}{50,100,180}
\usepackage[most]{tcolorbox}
\usepackage{listings}
\usepackage{graphicx}
\usepackage{subcaption}   % 提供子图环境
\usepackage[export]{adjustbox}  % 用于调整对齐
\newcommand{\gainstd}[3]{#1 ($\pm$ #2)\,\textcolor{green!45!black}{\tiny$\uparrow$#3}}
\newcommand{\gain}[2]{#1\,\textcolor{green!45!black}{\tiny$\uparrow$#2}}

\definecolor{takeawaybg}{HTML}{E8F4FD}      % 背景浅蓝
\definecolor{takeawayline}{HTML}{4472C4}    % 边框蓝

\newtcolorbox{takeawaybox}{
    enhanced,
    colback=takeawaybg,
    colframe=takeawayline,
    boxrule=0pt,          % 无边框
    leftrule=3pt,         % 左侧粗线
    arc=0pt,              % 无圆角
    outer arc=0pt,
    left=8pt,
    right=8pt,
    top=6pt,
    bottom=6pt,
    breakable,             % 允许跨页
    before skip=8pt,
    after skip=8pt,
}

\usepackage[utf8]{inputenc} % allow utf-8 input
\usepackage[T1]{fontenc}    % use 8-bit T1 fonts
\usepackage{hyperref}       % hyperlinks
\usepackage{url}            % simple URL typesetting
\usepackage{amsfonts}       % blackboard math symbols
\usepackage{microtype}      % microtypography
\usepackage{colortbl}
\title{%Auto-Rubric as Reward: Shaping Endogenous Criteria for Multimodal Generation
Auto-Rubric as Reward: From Implicit Preferences to Explicit Multimodal Generative Criteria
}

\author{%
  \textbf{Juanxi Tian}$^{1}$ \quad \textbf{Fengyuan Liu}$^{1}$ \quad \textbf{Jiaming Han}$^{3}$ \quad \textbf{Yilei Jiang}$^{3}$ \quad \\ 
  \textbf{Yongliang Wu}$^{4}$ \quad \textbf{Yesheng Liu}$^{1}$ \quad \textbf{Haodong Li}$^{1}$ \quad \textbf{Furong Xu}$^{2}$ \quad \\ 
  \textbf{Wanhua Li}$^{1,}$%
\setcounter{footnote}{1}%
\thanks{Corresponding author: \texttt{wanhua.li@ntu.edu.sg}} \vspace{0.4em} \\
  \normalfont $^1$Nanyang Technological University \quad 
  $^2$Ant Group \\ 
  $^3$MMLab, The Chinese University of Hong Kong \quad 
  $^4$UIUC \vspace{0.2em} \\ 
}
\begin{document}

\maketitle

\begin{abstract}
Aligning multimodal generative models with human preferences demands reward signals that respect the compositional, multi-dimensional structure of human judgment. Prevailing RLHF approaches reduce this structure to scalar or pairwise labels, collapsing nuanced preferences into opaque parametric proxies and exposing vulnerabilities to reward hacking. While recent Rubrics-as-Reward (RaR) methods attempt to recover this structure through explicit criteria, generating rubrics that are simultaneously reliable, scalable, and data-efficient remains an open problem. We introduce Auto-Rubric as Reward (ARR), a framework that reframes reward modeling from implicit weight optimization to explicit, criteria-based decomposition. Before any pairwise comparison, ARR externalizes a VLM's internalized preference knowledge as prompt-specific rubrics, translating holistic intent into independently verifiable quality dimensions. This conversion of implicit preference structure into inspectable, interpretable constraints substantially suppresses evaluation biases including positional bias, enabling both zero-shot deployment and few-shot conditioning on minimal supervision. To extend these gains into generative training, we propose Rubric Policy Optimization (RPO), which distills ARR's structured multi-dimensional evaluation into a robust binary reward, replacing opaque scalar regression with rubric-conditioned preference decisions that stabilize policy gradients. On text-to-image generation and image editing benchmarks, ARR-RPO outperforms pairwise reward models and VLM judges, demonstrating that explicitly externalizing implicit preference knowledge into structured rubrics achieves more reliable, data-efficient multimodal alignment, revealing that the bottleneck is the absence of a factorized interface, not a deficit of knowledge.
Code is publicly available at \url{https://github.com/OpenEnvision/AutoRubric-as-Reward}.
\end{abstract}

\section{Introduction}
\label{intro}
Human preferences are not arbitrary signals but structured, multidimensional judgments encompassing aesthetic value, semantic fidelity, and contextual appropriateness \cite{kirstain2023pick, xu2023imagereward, ma2025hpsv3}. Aligning generative multimodal models with such preferences therefore demands more than calibration: it requires models to internalize and operationalize the explicit criteria that underpin human evaluation. Prevailing RLHF paradigms contravene this requirement. By collapsing composite preference structures into scalar scores \cite{xu2023imagereward, ma2025hpsv3} or pairwise labels \cite{kirstain2023pick}, they encode rich human judgment into opaque, entangled representations, discarding the very dimensions that confer interpretability and stability, and exposing the learning process to reward hacking \cite{fan2023dpok, black2023training}.

Despite their extensive world knowledge and perceptual capabilities, contemporary VLMs exhibit systematic unreliability in modeling human preferences \cite{wang2024large, hu2025mmrewardbench2}. Pointwise scoring reduces evaluation to a single scalar, providing no constraint on how improvement is achieved and allowing degenerate optimization strategies. Pairwise comparison, while more balanced, still operates on a latent decision boundary, leading to persistent positional biases that resist standard mitigations such as positional labeling or chain-of-thought prompting \cite{wang2024large, liu2026examining}. Recent Rubrics as Reward (RaR) approaches attempt to recover structure through explicit criteria; however, their reliance on fixed or supervised rubric construction limits scalability, prompt specificity, and data efficiency, with these limitations becoming more pronounced when extended to multimodal generation settings.

The reframing recasts multimodal alignment as a representation problem: the bottleneck is not a deficit of preference knowledge, but the absence of a stable, factorized interface for applying it. Building on training-free rubric extraction from preference pairs~\cite{xie2025auto}, we propose \textbf{Auto-Rubric as Reward (ARR)}. ARR synthesizes instance-conditioned rubrics through a generate-verify-refine pipeline that induces discriminative criteria grounded in observable evidence, producing a compact set of verifiable, decision-relevant constraints spanning semantic fidelity, spatial consistency, compositional aesthetics, and edit faithfulness~\cite{lee2023holistic,ghosh2023geneval,zhang2023magicbrush,sheynin2024emu}. These criteria compose a structured evaluation protocol for criterion-level comparison, supplanting holistic scoring.
Unlike handcrafted rubrics or learned scalar rewards, ARR derives prompt-specific decision structures from minimal preference data with \emph{no parameter updates}, yielding a highly data-efficient and interpretable interface. By externalizing preference structure into explicit, verifiable criteria, ARR replaces unstable latent comparisons with grounded discrimination, helping to reduce positional bias and mitigating reward hacking. Crucially, rubric quality scales with the underlying VLM's alignment with human preferences: stronger judges produce more precise criteria without additional supervision.

This formulation extends from evaluation to optimization. If preference is inherently factorized, reward should preserve that structure rather than collapse it. We therefore introduce \textbf{Rubric Policy Optimization (RPO)}, which uses ARR-generated criteria to produce binary preference decisions for policy optimization. Unlike prior rubric-based methods that apply criteria as auxiliary filters, RPO integrates rubric-conditioned judgments directly into the optimization objective, aligning gradient updates with interpretable dimensions of quality. This eliminates a separate reward model and mitigates reward hacking by grounding supervision in explicit criteria rather than learned proxies \cite{black2023training, luo2025editscore}. Evaluation and generation are unified through a shared preference representation, where better understanding of human preferences in evaluation directly strengthens generative alignment.

Empirically, ARR improves preference accuracy over trained reward models and direct VLM judges by 1.7 to 6.3 points, while reducing positional bias and retaining strong zero-shot and few-shot generalization. When used for training, ARR-RPO yields further gains on text-to-image generation and image editing benchmarks~\cite{ma2025hpsv3, ghosh2023geneval, wu2025editreward, hu2024ella, hu2025mmrewardbench2, wei2025tiifbench, wang2026unigenbench, liu2025step1x, ye2025imgedit} (e.g., GenEval: 0.66 to 0.80; DPG-Bench: 83.84 to 85.76). These improvements require no judge fine-tuning or large-scale reward annotation. The core insight is that the bottleneck in multimodal alignment lies not in acquiring more preference knowledge, but in providing a stable, factorized interface to apply it, precisely what explicit rubrics supply.
%Empirically, ARR improves evaluation reliability over both pairwise reward models and direct VLM judges, significantly reducing positional bias while preserving strong zero-shot and few-shot generalization. When integrated into training, ARR-RPO translates these improvements into consistent gains in generative quality, outperforming existing reward modeling approaches on text-to-image generation and image editing benchmarks \cite{xu2023imagereward, ma2025hpsv3, wu2025editreward, ghosh2023geneval, lee2023holistic, zhang2023magicbrush, sheynin2024emu, li2026ueval}. These results point to a central conclusion: multimodal alignment is limited not by missing preference knowledge, but by the absence of a stable, factorized interface for applying it. Explicit rubric-based decomposition provides such an interface.

\noindent Our key contributions can be summarized as follows:
\begin{itemize}[leftmargin=*]

% \item \textbf{Auto-Rubric as Reward (ARR).} ARR externalizes latent human preferences into explicit prompt-conditioned rubrics, providing a robust, interpretable reward interface that mitigates positional bias and enables strong generalization under minimal supervision.

% \item \textbf{Rubric Policy Optimization (RPO).} RPO aggregates multiple rubric dimensions into a joint binary decision, integrating diverse preference signals within a contrastive framework rather than collapsing them into a single opaque scalar proxy.

% \item \textbf{Unified Evaluation-to-Optimization Framework.} By bridging evaluation and optimization via composable criteria, this paradigm replaces scalar rewards with factorized scoring, consistently enhancing generation quality across text-to-image and editing benchmarks.

\item \textbf{Auto-Rubric as Reward (ARR).} We propose a training-free framework that externalizes implicit human preferences into instance-conditioned, interpretable rubrics. It enables scalable multimodal evaluation with extremely high data efficiency, requiring only a few annotated samples.

\item \textbf{Rubric Policy Optimization (RPO).} We introduce RPO, a policy optimization framework for contrastive preference learning. By conditioning on ARR-derived rubrics, RPO replaces scalar reward signals with structured, criterion-grounded comparisons.
%%%%%%%%%%%%%%
\item \textbf{Diagnosing the Interface Bottleneck.} Ablations reveal the core bottleneck is a missing factorized interface, not a knowledge deficit. ARR-RPO resolves this via explicit rubrics; cross-model and cardinality analyses confirm that deeper comprehension of intrinsic criteria, rather than scale or data volume, drives both evaluation robustness and generative improvement.
%\item \textbf{Unified Evaluation-to-Optimization Framework.} ARR and RPO together establish an alignment paradigm that bridges evaluation and optimization through explicit, composable criteria. This unified framework yields consistent gains in both preference accuracy and generation quality over opaque scalar rewards, without additional supervision.

%ARR and RPO together establish a unified alignment paradigm that bridges evaluation and optimization via explicit, composable criteria. Structured, factorized scoring consistently outperforms opaque scalar rewards in generation quality.

%ARR and RPO together form a unified, score-centric multimodal alignment paradigm. Under this paradigm, preference modeling no longer relies on opaque scalar rewards but is instead based on explicit and composable scoring criteria, consistently translating these advantages into improved generation quality in text-to-image and image editing benchmarks.

\end{itemize}

\section{Related work}
\label{related}
\paragraph{Multimodal reward modeling.}
%RLHF underpins alignment across text-to-image, editing, and video generation. Early scalar reward models (PickScore, ImageReward, HPS) collapse preferences into scalar scores \cite{kirstain2023pick, xu2023imagereward, ma2025hpsv3}; while effective for coarse judgments, they invite reward hacking and overfitting due to opaque compression \cite{fan2023dpok, black2023training}. Direct preference optimization methods (Diffusion-DPO, D3PO) bypass explicit reward models but still reduce rich preferences to scalar or pairwise signals, risking spurious correlations \cite{fan2023dpok, black2023training, zheng2025diffusionnft}. VLM-as-a-judge approaches leverage richer world knowledge yet exhibit persistent positional, symmetry, and visual biases that resist standard mitigations \cite{wang2024large, liu2026examining, hu2025mmrewardbench2, zhao2026trust}. We contend the bottleneck is not insufficient preference knowledge, but the absence of a stable, factorized interface. By externalizing a VLM's implicit priors into explicit, prompt-conditioned rubrics \cite{xie2025auto}, our method decomposes holistic judgment into independently verifiable quality axes, replacing lossy scalar compression with structured discrimination.
RLHF underpins alignment across text-to-image generation, editing, and video synthesis.
Early reward models such as PickScore, ImageReward, and HPS compress rich human preferences into scalar signals \cite{kirstain2023pick, xu2023imagereward, ma2025hpsv3}. While effective for coarse ranking, such compression obscures preference structure and is prone to reward hacking and overfitting \cite{black2023training,zheng2025diffusionnft}.
Direct optimization methods eliminate explicit reward modeling but still rely on scalar or pairwise objectives, inheriting similar limitations in expressivity and robustness \cite{fan2023dpok,wallace2023diffusiondpo}.
Recent VLM-as-a-judge approaches leverage stronger multimodal priors, yet exhibit persistent biases, such as positional and symmetry bias, that are difficult to eliminate through prompting alone \cite{wang2024large, liu2026examining, hu2025mmrewardbench2, zhao2026trust}.
Taken together, these methods suggest that the core limitation is not a lack of preference knowledge, but the absence of a structured interface for expressing and applying it.
We address this by externalizing implicit preferences into explicit, prompt-conditioned rubrics, enabling factorized and verifiable evaluation in place of opaque scalar scoring.

\paragraph{Rubric as reward.}
To overcome the limitations of scalar evaluation, recent work has explored rubric-based formulations that decompose judgments into interpretable criteria. In language tasks, analytic rubric frameworks \cite{pathak2025rubric, ye2023flask} and LLM-Rubric \cite{hashemi2024llmrubric} show that criterion-level assessment yields more stable and calibrated signals than holistic scoring \cite{kim2023prometheus, ankner2024critique, pan2026rubriceval}. AutoRubric \cite{xie2025auto} extends this idea by distilling generalizable criteria from preference data, yet remains confined to text-only evaluation. In multimodal settings, AutoRubric-R1V \cite{jia2026autorubricrubricbasedgenerativerewards} compiles consistent reasoning steps from successful trajectories into problem-specific rubrics for process-level supervision, but it is designed for vision-language reasoning, not generative policy optimization. Despite these advances, no prior method in multimodal generation adopts auto-generated rubrics as the reward for both evaluation and training\cite{zhao2026trust,li2026hpedit}. We address this gap by treating rubrics as the direct preference interface, instantiating them as explicit, prompt-conditioned criteria that govern evaluation and provide the reward signal for optimization. This reframes alignment from implicit scalar optimization to structured discrimination over verifiable criteria, yielding a more interpretable and robust reward.

\section{Methodology}
\label{method}
\begin{figure}[H]
    \centering
    \makebox[\textwidth][c]{%
        \begin{minipage}{1.0\textwidth}
            \centering
            \includegraphics[width=\linewidth]{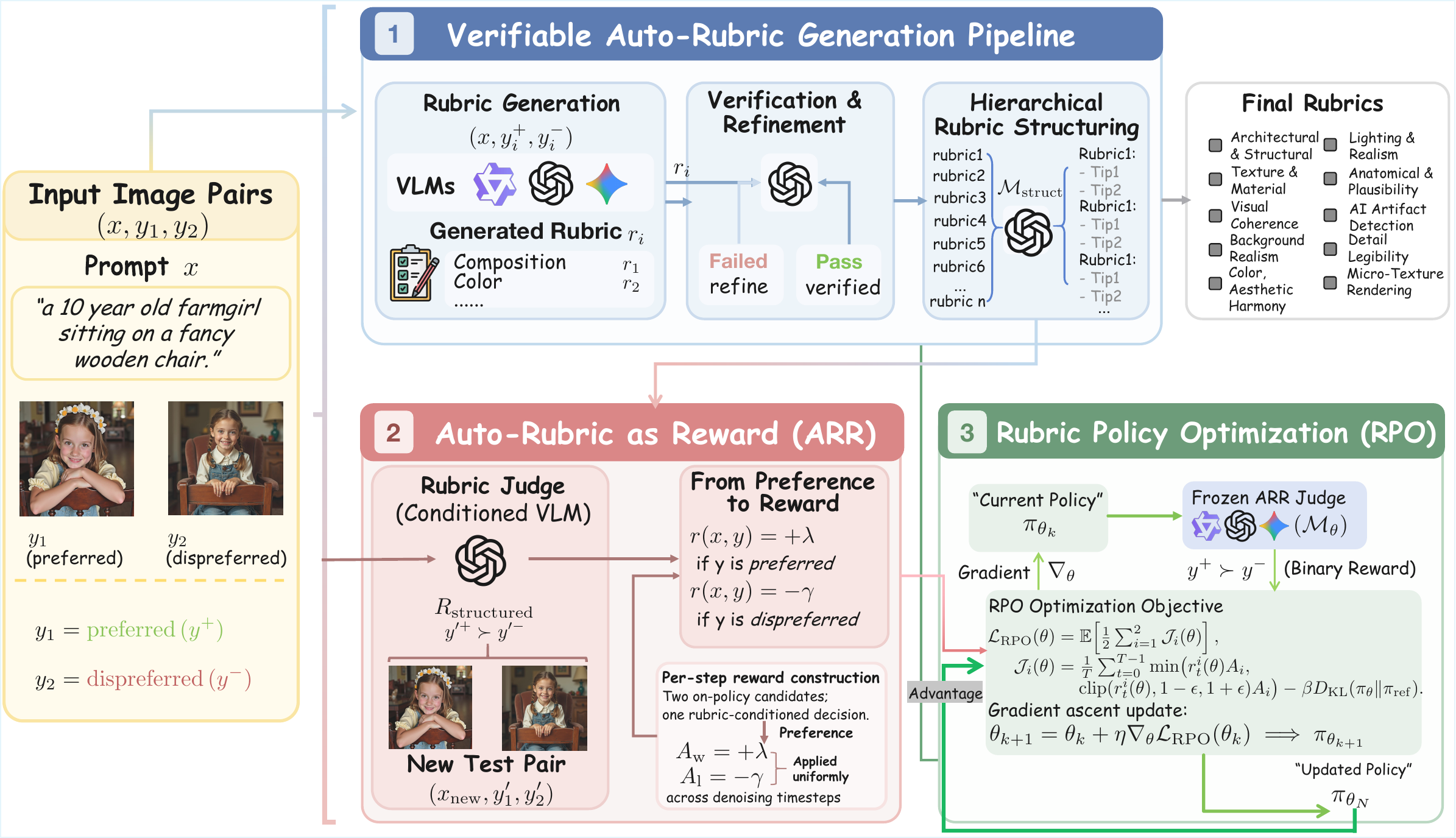}
            \caption{\textbf{Overview of the ARR-RPO framework.}(1) Rubrics are generated from preference pairs, verified, and hierarchically structured. (2) ARR converts rubric-conditioned VLM judgments into binary rewards. (3) RPO uses these rewards from a frozen judge to optimize the generative policy.}
            \label{fig:methods}
        \end{minipage}%
    }
\end{figure}
\subsection{Problem formulation}
\label{subsec:formulation}
We formulate preference learning as estimating the optimal parameters of a probabilistic model $P_{\theta}$ that, given a prompt $x$ and candidate outputs $y^+, y^-$, assigns higher likelihood to the response better satisfying human intent. Preference alignment thus optimizes $P_{\theta}$ to capture and generalize human preferences, raising the central design question: how should the parameters $\theta$ be specified? We address this by decomposing the problem into ARR for evaluation and RPO for training (Figure~\ref{fig:methods}).
%We formulate preference learning as finding the optimal parameters of a probabilistic preference model $  P_{\theta}  $. Let $  P_{\theta}  $ denote a human preference model that, given a prompt $  c  $ and a pair of outputs $  (x^+, x^-)  $, assigns higher likelihood to the response that better satisfies human intent. The goal of preference alignment is to optimize $  P_{\theta}  $ such that it accurately captures and generalizes human preferences. The core design question is: how should the parameters $  \theta  $ that define this preference model be specified?

\paragraph{Implicit preference modeling.}
For implicit preference modeling, given a pair of outputs $  (y^+, y^-)  $ conditioned on the same input $  x  $, the human preference probability is typically defined using the Bradley-Terry (BT) model as follows:

\begin{equation}
P^*(y^+ \succ y^- \mid x) = \frac{\exp(r^*(x, y^+))}{\exp(r^*(x, y^+)) + \exp(r^*(x, y^-))}
\label{eq:implicit_bt}
\end{equation}

where $  *  $ denotes the parameters corresponding to the true underlying human preference distribution. Here, $  r^*  $ represents the ideal scalar reward model that perfectly reflects human preferences. In practice, since the true human preference distribution is inaccessible, we typically work with a pairwise preference dataset $  \mathcal{D}  $ that approximately captures human judgments. We can then parameterize a reward model $r_\phi$ and estimate the true parameters $\phi^*$ by solving the following optimization problem:
\begin{equation}
\mathcal{L}_R(r_\phi, \mathcal{D}) = -\mathbb{E}_{(x,y^+,y^-) \sim \mathcal{D}} \left[ \log \sigma(r_\phi(x, y^+) - r_\phi(x, y^-)) \right]
\label{eq:implicit_loss}
\end{equation}
where $\sigma$ is the logistic function.

% While effective, this approach produces an opaque black-box model with limited interpretability. By collapsing multi-dimensional human preferences into a single scalar score, it marginalizes the rich structure of human judgment and creates an underdetermined training signal. Moreover, optimizing against a fixed parametric reward model $  r_{\phi}  $ inevitably leads to reward hacking, as the generator exploits spurious correlations in the reward rather than truly satisfying human intent, which is a direct consequence of Goodhart's Law that worsens as the policy diverges from the true human preference distribution.

\paragraph{Explicit preference modeling.}
In explicit preference modeling, we define the preference distribution by employing a VLM as a judge. Given a paired input $  (x, y^{+}, y^{-})  $, the LLM judge processes the prompt $  x  $ along with the two candidate outputs and produces a binary preference decision that approximates the underlying human preference distribution $  P_{\theta}  $:
\begin{equation}
P_{\theta}(y^{+} \succ y^{-} | x) = \mathcal{M}_{\theta}(y^{+} \succ y^{-} \mid x, y^{+}, y^{-}, R),
\label{eq:explicit}
\end{equation}

where $  R  $ is a carefully pre-defined natural language rubric designed to enhance the VLM's ability to discern subtle differences in response quality. Here, $  \mathcal{M}_{\theta}  $ denotes the VLM enhanced by $  R  $, which serves as the judge and outputs a binary preference decision between the two candidates.

% \begin{table}[t]
% \centering
% \footnotesize
% \setlength{\tabcolsep}{5pt}
% \renewcommand{\arraystretch}{1.15}

% \caption{Paradigm shift from implicit to explicit reward parameterization.}

% \begin{tabular}{lccccc}
% \toprule
% \textbf{Method} & \textbf{Reward Form} & \textbf{Reward Hacking} & \textbf{Interpretability} & \textbf{Data Scale} & \textbf{Training Overhead} \\
% \midrule
% Pointwise RM & Implicit scalar & High  & Low & Large & High \\
% Pairwise RM & Implicit comparative & Medium & Low & Large & High \\
% VLM-as-Judge & Implicit reasoning & Medium--High & Low & -- & -- \\
% \midrule
% \rowcolor{cyan!5}
% \textbf{ARR (Ours)} & Explicit rubric & Low & High & Minimal & -- \\
% \bottomrule
% \end{tabular}
% \end{table}

\subsection{Auto-Rubric as Reward}
\label{subsec:arr}

Let $\mathcal{S}$ be the space of all possible rubrics. We aim to find the optimal rubric $R^*$ that best approximates the underlying human preference distribution. Given an ideal preference model $P^*$ instantiated by a highly capable LLM judge, the optimal rubric can be formulated as:

%\begin{equation}
%R^{*} = \arg\max_{R \subset \mathcal{S}} \sum_{i=1}^{N} \log P^*(y_{i}^{+} \succ y_{i}^{-} | x_{i})
%\label{eq:rubric_gen}
%\end{equation}

\begin{equation}
R^{*} = \arg\max_{R \subset \mathcal{S}} \sum_{i=1}^{N} \log P^*(y_{i}^{+} \succ y_{i}^{-} | x_{i}, R)
\label{eq:rubric_gen}
\end{equation}

Since the space of all possible rubric sets $\mathcal{S}$ is vast and discrete, directly optimizing the ideal objective is intractable. We therefore simplify the optimization target as selecting the best rubric subset:
\begin{equation}
R^{*} \approx \arg\max_{R \subset \mathcal{D}_R} \sum_{i=1}^{N} \mathbb{I}[\mathcal{M}_{\theta}(y_i^{+} \succ y_i^{-} \mid x_i, y_i^{+}, y_i^{-}, R) = \text{correct}],
\label{eq:approximate_objective}
\end{equation}
%\begin{equation}
%R^{*} \approx \arg\max_{R \subset \mathcal{D}_R} \sum_{i=1}^{N} \mathbb{I}[\mathcal{M}_{\theta}(x_i, y_i^{+}, y_i^{-}\mid x_i, y_i^{+}, y_i^{-}, R) = \text{correct}],
%\label{eq:approximate_objective}
%\end{equation}

where $  \mathcal{D}_R  $ is a finite set of candidate rubrics. In the remainder of this section, we detail our approach for automatically constructing high-quality rubrics from data and demonstrate how these auto-generated rubrics can serve as an interpretable and effective reward signal when applied to reinforcement learning tasks.

\paragraph{Verifiable rubric generation.}

Given a pairwise preference dataset \(\mathcal{D} = \{(x_i, y_i^+, y_i^-)\}_{i=1}^N\), we first generate \(K\) candidate rubrics per labeled pair offline, with \(K=5\) by default.
For every pair \((x_i, y_i^+, y_i^-)\), an VLM is prompted to produce a detailed natural language rubric \(r_i\) that explains why \(y_i^+\) is preferred over \(y_i^-\):
\begin{equation}
r_i = \mathcal{M}_{\text{gen}}(x_i, y_i^+, y_i^-).
\end{equation}

To ensure quality, each generated rubric \(r_i\) is then verified by a separate judgment step. The verifier checks whether the rubric consistently supports the original preference:
\begin{equation}
v_i = \mathcal{M}_{\text{verify}}(x_i, y_i^+, y_i^-, r_i).
\end{equation}

Because the verifier independently checks whether the generated rubric consistently recovers the original preference label, it acts as a weak safeguard against self-reinforcing errors: rubrics that fail this consistency test are refined or discarded, reducing the chance of amplifying idiosyncratic model biases that survive the initial generation step.

If verification fails (\(v_i = \text{false}\)), we iteratively refine the rubric up to a predefined maximum number of attempts \(T_{\max}\):
\begin{equation}
r_i^{(t+1)} = \mathcal{M}_{\text{refine}}(x_i, y_i^+, y_i^-, r_i^{(t)}), \quad t = 0, 1, \dots, T_{\max}-1.
\end{equation}

If the rubric still fails verification after \(T_{\max}\) refinement attempts, it is discarded. After processing all \(K\) candidates for each pair in \(\mathcal{D}\), we obtain a set of verified rubrics:
\begin{equation}
\mathcal{D}_R = \{ r_{i,k} \mid v_{i,k} = \text{true} \}.
\end{equation}

This verifiable generation process yields a high-quality, instance-specific rubric collection \(\mathcal{D}_R\) directly grounded in the preference dataset.

\paragraph{Hierarchical rubric structuring.}
After verification, the rubric set $\mathcal{D}_R$ captures fine-grained, per-instance criteria but lacks the coherence required for consistent conditioning across arbitrary prompts. We therefore prompt an LLM to consolidate $\mathcal{D}_R$ into a single, hierarchically organized rubric. The LLM groups related criteria by semantic granularity and preference dimension, producing a compact evaluation protocol. The resulting structured rubric $R_{\text{structured}}$ is directly reused as a system-prompt component for the judge and as a reward conditioning signal during optimization, removing the need for per-instance rubric regeneration at deployment.
Formally,
\begin{equation}
R_{\text{structured}} = \mathcal{M}_{\text{struct}}(\mathcal{D}_R),
\end{equation}
where $\mathcal{M}_{\text{struct}}$ denotes the LLM prompted to perform hierarchical organization and prompt synthesis. See Appendix~\ref{app:example_rubrics} for final rubric examples.
%For the final output, we employ an LLM to organize the rubric set $  \mathcal{D}_R $ into a well-structured, hierarchical prompt. Specifically, the model synthesizes the collected rubrics into a coherent hierarchical structure that reflects different levels of semantic granularity and preference criteria. This process produces the final optimized rubric set $  R_{\text{structured}}  $, which can be directly used as a conditioning signal for the judge model or a reward model in reinforcement learning.
%Formally, the hierarchical structuring step is defined as:
%\begin{equation}
%R_{\text{structured}} = \mathcal{M}_{\text{struct}}\Bigl(\mathcal{D}_R\Bigr)
%\end{equation}

%where $  \mathcal{M}_{\text{struct}}  $ denotes the large language model prompted to perform hierarchical organization and prompt synthesis.

\paragraph{From rubric to reward.}

To successfully apply the auto-rubric method to reinforcement learning tasks, we need to convert the generated rubrics into a usable reward signal. Since the VLM judge produces binary preference decisions, we assign a positive constant reward to the preferred response $y^+$ and a negative constant reward to the dispreferred response $y^-$. Formally, given a prompt $x$ and a pair of outputs $(y^+, y^-)$, the reward for a candidate $y$ is defined with respect to the other output $y'$ as:
\begin{equation}
r(x, y; y') = 
\begin{cases} 
+ \lambda & \text{if } \mathcal{M}_{\theta}(x, y, y', R) \text{ prefers } y, \\
- \gamma & \text{otherwise},
\end{cases}
\end{equation}
where $\lambda, \gamma > 0$ are constant reward magnitudes and $R$ denotes the learned rubric set.

\subsection{Rubric Policy Optimization}
\label{subsec:rpo}

Having established a mechanism for generating high-quality rubrics and converting them into verifiable reward signals, we now introduce \textbf{Rubric Policy Optimization (RPO)}, an online policy optimization algorithm that directly utilizes the rubric judge to guide the generative policy \(\pi_{\theta}\).

%Unlike conventional RLHF that relies on scalar rewards, RPO directly uses the VLM judge's binary preference decisions as the reward signal. 
Unlike conventional RLHF and prior rubric-based methods in multimodal generation that reduce criteria to scalar composites or auxiliary filters, RPO directly leverages the VLM judge's binary preferences conditioned on explicit rubrics as the reward signal. For each generated sample, the preferred output \(y^+\) receives a positive constant reward \(+\lambda\), while the dispreferred output \(y^-\) receives \(-\gamma\). This yields a dense per-step training objective that preserves the advantages of rubric-based evaluation while remaining compatible with standard policy gradient methods.

The resulting RPO objective is defined as:
\begin{equation}
\begin{aligned}
\mathcal{L}_{\mathrm{RPO}}(\theta) 
&= \mathbb{E}_{h \sim \mathcal{D},\, \{x^i_{0:T}\}_{i=1}^2 \sim \pi_{\theta}} \Bigg[ 
\frac{1}{2} \sum_{i=1}^{2} \Bigg( 
\frac{1}{T} \sum_{t=0}^{T-1} \min\Bigl( r_t^i(\theta) A_i, \\
&\quad \text{clip}(r_t^i(\theta), 1-\epsilon, 1+\epsilon) A_i \Bigr) 
- \beta \, D_{\mathrm{KL}}(\pi_\theta \| \pi_{\mathrm{ref}})
\Bigg) 
\Bigg].
\end{aligned}
\label{eq:rpo_objective}
\end{equation}
where the importance ratio at each timestep is
\begin{equation}
r_t^i(\theta)= \frac{\pi_{\theta}(x_{t-1}^i \mid x_t^i, h)}{\pi_{\theta_{\mathrm{old}}}(x_{t-1}^i \mid x_t^i, h)}.
\end{equation}

\paragraph{Per-step reward construction.}
For a given prompt \(h\) (which may include both text condition \(c\) and the current rubric \(R\)), we sample two trajectories from the current policy \(\pi_\theta\). The VLM judge, conditioned on the learned rubric, produces a binary preference decision between the two trajectories. The winning trajectory is assigned advantage \(A_w = +\lambda\) and the losing one \(A_l = -\gamma\). This per-trajectory advantage is then uniformly distributed across all denoising (or generation) timesteps, providing a dense training signal that directly reflects rubric-guided human preference.

\paragraph{Online optimization and robustness.}
RPO optimizes the policy online: each iteration samples prompts from $\mathcal{D}$, generates two candidates from $\pi_{\theta}$, evaluates them in one call to the frozen rubric judge using the cached $R_{\text{structured}}$
, and applies the gradient of $\mathcal{L}_{\mathrm{RPO}}(\theta)$. Because rewards come from a frozen VLM judge conditioned on explicit rubrics rather than a trainable scalar model, RPO helps mitigate reward hacking. As the policy evolves, the judge continues to apply the same task rubric. PPO-style clipping and KL regularization further stabilize training and enable exploration aligned with the multi-dimensional criteria in the rubrics. 
%This factorized reward mechanism yields more stable optimization and stronger performance than collapsing human preferences into a single scalar.
%RPO operates fully online: each iteration samples prompts from $\mathcal{D}$, generates two candidates from $\pi_{\theta}$, evaluates them via the rubric judge, and applies the gradient of $\mathcal{L}_{\mathrm{RPO}}(\theta)$. Because rewards derive from a frozen (M)LLM judge conditioned on explicit rubrics rather than a trainable scalar model, RPO substantially mitigates reward hacking. Rubrics are dynamically regenerated for each prompt and policy output pair, so the optimization target adapts naturally to the evolving distribution of $\pi_{\theta}$, conferring robustness against distributional shift. PPO-style clipping and KL regularization further stabilize training, enabling genuine exploration aligned with the multi-dimensional criteria encoded in the rubrics. This factorized reward mechanism achieves more stable optimization and stronger performance than methods that collapse human preferences into a single scalar signal.

\section{Experiments}
\label{experiments}
We evaluate \textsc{ARR} as a preference evaluator and as a structured reward for generative policy optimization. Experiments on multimodal understanding, text-to-image generation, and image editing benchmarks compare against trained reward models and direct VLM judges to assess gains in evaluative reliability and downstream performance.
%We evaluate \textsc{ARR} in two roles: as an evaluator of preference judgments and as a structured reward for generative policy optimization (\textsc{RPO}). Experiments are conducted on both multimodal understanding, text-to-image generation and image editing benchmarks, comparing against strong trained reward models and direct VLM-as-Judge baselines to assess gains in evaluative reliability and downstream performance.
\subsection{Experimental setup}
\label{subsec:setup}
\begin{table}[t]
\centering
\footnotesize
\setlength{\tabcolsep}{3.5pt}
\renewcommand{\arraystretch}{1.12}
\caption{\textbf{Evaluator performance across four preference benchmarks.} Accuracy (\%) denotes agreement with human preference labels. The best result in each column is \textbf{bold}. Blue-shaded rows indicate ARR; green values indicate absolute gains over the corresponding baseline VLM judge.}
\label{tab:evaluator_main}
\begin{adjustbox}{max width=\textwidth}
\begin{tabular}{lcccc}
\toprule
\multirow{2}{*}{Method} & HPDv3 & \shortstack{MM-RewardBench2\\(T2I)} & \shortstack{MM-RewardBench2\\(Edit)} & EditReward-Bench \\
& Acc. & Acc. & Acc. & Acc. \\
\midrule

\multicolumn{5}{c}{\textbf{\textit{Trained Reward Model}}} \\
\midrule
PickScore & 65.6 & 58.6 & --- & --- \\
ImageReward & 58.6 & 54.0 & --- & --- \\
UnifiedReward & 66.0 & 59.8 & --- & --- \\
UnifiedReward-Thinking & 68.1 & 66.0 & --- & --- \\
HPSv3 & 76.9 & 60.2 & --- & --- \\
EditReward & --- & --- & 67.2 & 56.45 \\
\midrule

\multicolumn{5}{c}{\textbf{\textit{VLM-as-Judge w/o ARR}}} \\
\midrule

Qwen3-VL-8B & 67.2 & 57.6 & 59.2 & 54.01 \\
\rowcolor{cyan!5} w/ ARR & \gainstd{70.2}{0.2}{3.0} & \gainstd{62.7}{0.2}{5.1} & \gainstd{65.5}{0.3}{6.3} & \gainstd{57.22}{0.1}{3.21} \\
\midrule

GPT-5 & 72.4 & 70.5 & 73.8 & 57.53 \\
\rowcolor{cyan!5} w/ ARR & \gainstd{76.1}{0.2}{3.7} & \gainstd{74.7}{0.4}{4.2} & \gainstd{77.5}{0.3}{3.7} & \gainstd{61.01}{0.1}{3.48} \\
\midrule

Gemini~3.1~Pro & 76.6 & 75.1 & 77.4 & 61.23 \\
\rowcolor{cyan!5} w/ ARR & \gainstd{\textbf{78.3}}{0.1}{1.7} & \gainstd{\textbf{78.9}}{0.2}{3.8} & \gainstd{\textbf{79.2}}{0.2}{1.8} & \gainstd{\textbf{63.27}}{0.2}{2.04} \\

\bottomrule
\end{tabular}
\end{adjustbox}
\end{table}
\paragraph{Evaluation benchmarks.}
Evaluator fidelity is measured on three established testbeds: MM-RewardBench2 \cite{hu2025mmrewardbench2}, which provides fine-grained diagnostic splits across multimodal reward scenarios; HPDv3 (test set) \cite{ma2025hpsv3}, a large-scale text-to-image preference corpus comprising 14,400 pairwise human judgments; and EditReward-Bench \cite{wu2025editreward}, specifically curated to probe instruction adherence in image editing. For generative quality assessment, we adopt GenEval \cite{ghosh2023geneval}, DPG-Bench \cite{hu2024ella}, TIIF(test-mini-short)\cite{wei2025tiifbench}, and UniGenBench++\cite{wang2026unigenbench} for text-to-image synthesis, complemented by GEdit-Bench\cite{liu2025step1x} and ImgEdit\cite{ye2025imgedit} for editing tasks.

\paragraph{Baselines and implementation.}
For human preference evaluation, we compare against a suite of state-of-the-art trained reward models, including HPSv3 \cite{ma2025hpsv3}, PickScore \cite{kirstain2023pick}, ImageReward \cite{xu2023imagereward}, UnifiedReward\cite{wang2025unified} and UnifiedReward-Thinking \cite{wang2025unirewardthinking}, and EditReward \cite{wu2025editreward}, alongside representative VLM judges such as Qwen3-VL \cite{bai2025qwen3}, GPT-5 \cite{singh2025openai}, and Gemini~3.1~Pro \cite{deepmind2026gemini31pro}. 

Following the common practice in recent multimodal alignment and generation research \cite{hu2025mmrewardbench2,wallace2023diffusiondpo,li2026hpedit}, we adopt FLUX.1-dev \cite{labs2025flux} and Qwen-Image-Edit-2509 \cite{wu2025qwen-image} as base models for image generation and editing, respectively. We perform post-training with RPO on LoRA-adapted versions of these models. Training prompts are drawn from ShareGPT-4o-Image \cite{chen2025sharegpt}.
Unless otherwise specified, ARR generates five candidate rubrics per labeled training pair offline and consolidates the verified candidates into one cached rubric per task and experimental configuration. We further contextualize results against leading contemporary generative models.

%\paragraph{Baselines.}
%For evaluator benchmarking, we compare against a suite of state-of-the-art trained reward models comprising HPSv3 \cite{ma2025hpsv3}, PickScore \cite{kirstain2023pick}, ImageReward \cite{xu2023imagereward}, UnifiedReward \cite{wang2025unified}, UnifiedReward-Thinking \cite{wang2025unified} and EditReward \cite{wu2025editreward}, alongside representative VLM judges including Qwen3-VL \cite{bai2025qwen3}, GPT-5 \cite{singh2025openai}, and Gemini~3.1~Pro\cite{deepmind2026gemini31pro}. In generative experiments, RPO is evaluated relative to the base architectures FLUX.1.dev \cite{labs2025flux} and Qwen-Image-Edit-2509 \cite{wu2025qwen-image}, with further contextualization against leading contemporary models.

%\paragraph{ARR instantiation.}
%Unless noted otherwise, \textsc{ARR} prompts a frozen VLM to dynamically synthesize five prompt-conditional rubrics per instance, which subsequently score all candidate images. The variant ARR~(w/ guide) augments the generation meta-prompt with a minimal set of human preference exemplars, demonstrating that even sparse guidance materially sharpens rubric relevance.

%\paragraph{Training Setup.}
%The training prompts for both the editing and generation reinforcement learning processes are curated from the ShareGPT-4o-Image dataset \cite{chen2025sharegpt}. Both FLUX.1-dev \cite{labs2025flux} and Qwen-Image-Edit-2509 \cite{wu2025qwen-image} are fine-tuned using the LoRA (Low-Rank Adaptation) technique, which enables parameter-efficient adaptation while maintaining the original model weights.

\subsection{Human preference quality}
\label{subsec:evaluator_results}

We evaluate ARR as a preference evaluator on three standard benchmarks: HPDV3\cite{ma2025hpsv3}, which provides 1.17M human pairwise comparisons for text-to-image; MM-RewardBench2\cite{hu2025mmrewardbench2}, with 4,000 expert-annotated preference pairs spanning four tasks; and EditReward-Bench, covering 13 subtasks of instruction-guided editing. For each benchmark, we report pairwise preference accuracy, defined as the fraction of test pairs where the model's predicted preference matches the human judgment. For each ARR configuration, we report the mean and standard deviation of the accuracies from three independent evaluation runs.

\noindent\textbf{Results.}
Table~\ref{tab:evaluator_main} reports preference accuracy.
Pairwise reward models specialize narrowly (e.g., HPSv3 drops from 76.9\% on HPDv3 to 60.2\% on MM-RewardBench2 T2I; EditReward falls from 67.2\% to 56.5\% on the broader EditReward-Bench), while direct VLM judges generalize better yet still struggle on challenging splits (Gemini~3.1~Pro: 75.1--77.4\% on the first three columns but only 61.2\% on EditReward-Bench).
ARR conditioning consistently improves all judges by 1.7--6.3 points, with Gemini~3.1~Pro + ARR reaching state-of-the-art on all four benchmarks.
Critically, base VLMs exhibit severe positional bias ($\Delta = 30.2$--$34.6$; Table~\ref{tab:position_bias}); ARR reduces this gap to 27.8--31.6 (zero-shot) and to 8.9--10.3 with guidance.
Gains persist across model families (Table~\ref{tab:cross_model}), confirming that rubric quality, not generator-judge co-adaptation, drives results. Full results are in Table~\ref{tab:evaluator_app}.

\begin{takeawaybox}
\noindent\textbf{Takeaway:} Rubric conditioning does not merely improve accuracy; it reframes evaluation from preference matching to criteria-aligned verification. By externalizing standards \emph{prior} to comparison, ARR mitigates latent biases in implicit preference models, yielding structured, task-adaptive decisions that directly support robust reward modeling.
%Rubric conditioning resolves the tension between specialization and generalization by forcing evaluators to externalize criteria \emph{prior} to judgment, inducing task-adaptive behavior without parameter updates. 
%Rubric conditioning resolves the tension between specialization and generalization by forcing evaluators to externalize criteria \emph{prior} to judgment, inducing task-adaptive behavior without parameter updates. The resulting rubric functions as a portable, architecture-agnostic regularizer and a lightweight recalibration channel. This points to a central insight: the primary limitation in multimodal preference evaluation is not model capacity, but the absence of a stable interface for expressing latent preference knowledge, an interface naturally realized through explicit rubric generation.
\end{takeawaybox}

\subsection{Image generation and editing performance}
\label{subsec:generative_results}

\begin{table}[t]
\centering
\footnotesize
\setlength{\tabcolsep}{3.5pt}
\renewcommand{\arraystretch}{1.15}
\caption{\textbf{Generative performance across T2I and image editing benchmarks.}  Blue-shaded rows mark ARR-RPO variants; green values indicate absolute gains over the corresponding baseline.}
\label{tab:generative_main}
\begin{adjustbox}{max width=\textwidth}
\begin{tabular}{lccccccc}
\toprule
\multirow{2}{*}{Method} & \multicolumn{5}{c}{Text-to-Image} & \multicolumn{2}{c}{Image Editing} \\
\cmidrule(lr){2-6} \cmidrule(lr){7-8}
& GenEval & DPG-Bench & TIIF & \multicolumn{2}{c}{UniGenBench++} & GEdit-Bench & ImgEdit \\
\cmidrule(lr){5-6}
& & & & Short & Long & & \\
\midrule

\multicolumn{8}{c}{\textbf{\textit{Specialist Model (T2I)}}} \\
\midrule
Emu3 & 0.54 & 80.60 & --- & 45.42 & 50.59 & --- & --- \\
JanusFlow & 0.63 & 79.68 & --- & 47.10 & 54.80 & --- & --- \\
FLUX.1-Dev & 0.66 & 83.84 & 71.09 & 60.97 & 69.42 & --- & --- \\
DALLE-3 & 0.67 & 83.50 & 74.96 & 68.85 & 70.82 & --- & --- \\
Show-o2 & 0.76 & 86.14 & --- & 61.90 & 70.33 & --- & --- \\
OmniGen2 & 0.80 & 83.57 & --- & 63.09 & 71.39 & --- & --- \\
BAGEL & 0.82 & 85.07 & 71.50 & 59.91 & 71.26 & --- & --- \\
\midrule

\multicolumn{8}{c}{\textbf{\textit{ARR-RPO / T2I (Ours)}}} \\
\midrule
\rowcolor{cyan!5} w/ RPO-Qwen3vl-8B-ARR & \gain{0.74}{0.08} & \gain{85.03}{1.19} & \gain{74.92}{3.83} & \gain{64.17}{3.20} & \gain{71.82}{2.40} & --- & --- \\
\rowcolor{cyan!5} w/ RPO-GPT-5-ARR & \gain{0.78}{0.12} & \gain{85.41}{1.57} & \gain{76.18}{5.09} & \gain{65.36}{4.39} & \gain{72.41}{2.99} & --- & --- \\
\rowcolor{cyan!5} w/ RPO-Gemini 3.1 Pro-ARR & \gain{0.80}{0.14} & \gain{85.76}{1.92} & \gain{76.85}{5.76} & \gain{65.89}{4.92} & \gain{72.93}{3.51} & --- & --- \\
\midrule

\multicolumn{8}{c}{\textbf{\textit{Specialist Model (Image Editing)}}} \\
\midrule
Instruct-Pix2Pix & --- & --- & --- & --- & --- & 3.68 & 1.88 \\
AnyEdit & --- & --- & --- & --- & --- & 3.21 & 2.45 \\
Step1X-Edit & --- & --- & --- & --- & --- & 6.97 & 3.06 \\
Qwen-Image-Edit-2509 & --- & --- & --- & --- & --- & 7.54 & 4.35 \\
UniWorld-Qwen-Image-Edit & --- & --- & --- & --- & --- & 7.76 & 4.48 \\
\midrule

\multicolumn{8}{c}{\textbf{\textit{ARR-RPO / Image Editing (Ours)}}} \\
\midrule
\rowcolor{cyan!5} w/ RPO-Qwen3vl-8B-ARR & --- & --- & --- & --- & --- & \gain{7.66}{0.12} & \gain{4.38}{0.03} \\
\rowcolor{cyan!5} w/ RPO-GPT-5-ARR & --- & --- & --- & --- & --- & \gain{7.72}{0.18} & \gain{4.40}{0.05} \\
\rowcolor{cyan!5} w/ RPO-Gemini 3.1 Pro-ARR & --- & --- & --- & --- & --- & \gain{7.85}{0.31} & \gain{4.43}{0.08} \\
\bottomrule
\end{tabular}
\end{adjustbox}
\end{table}
\begin{figure}[t]
    \centering
    \makebox[\textwidth][c]{%
        \begin{minipage}{\textwidth}
            \centering
            \includegraphics[width=\linewidth]{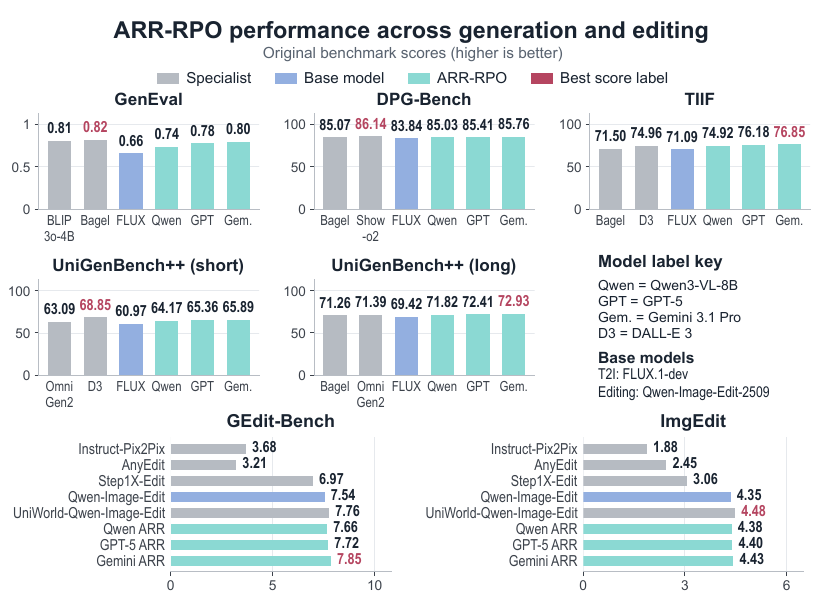}
            \caption{\textbf{Performance comparison of ARR-RPO variants against specialist models} across text-to-image generation (top) and image editing (bottom) benchmarks.}
            \label{fig:performance}
        \end{minipage}%
    }
\end{figure}

\begin{figure}[t]
    \centering
    \makebox[\textwidth][c]{%
        \begin{minipage}{\textwidth}
            \centering
            \includegraphics[width=\linewidth]{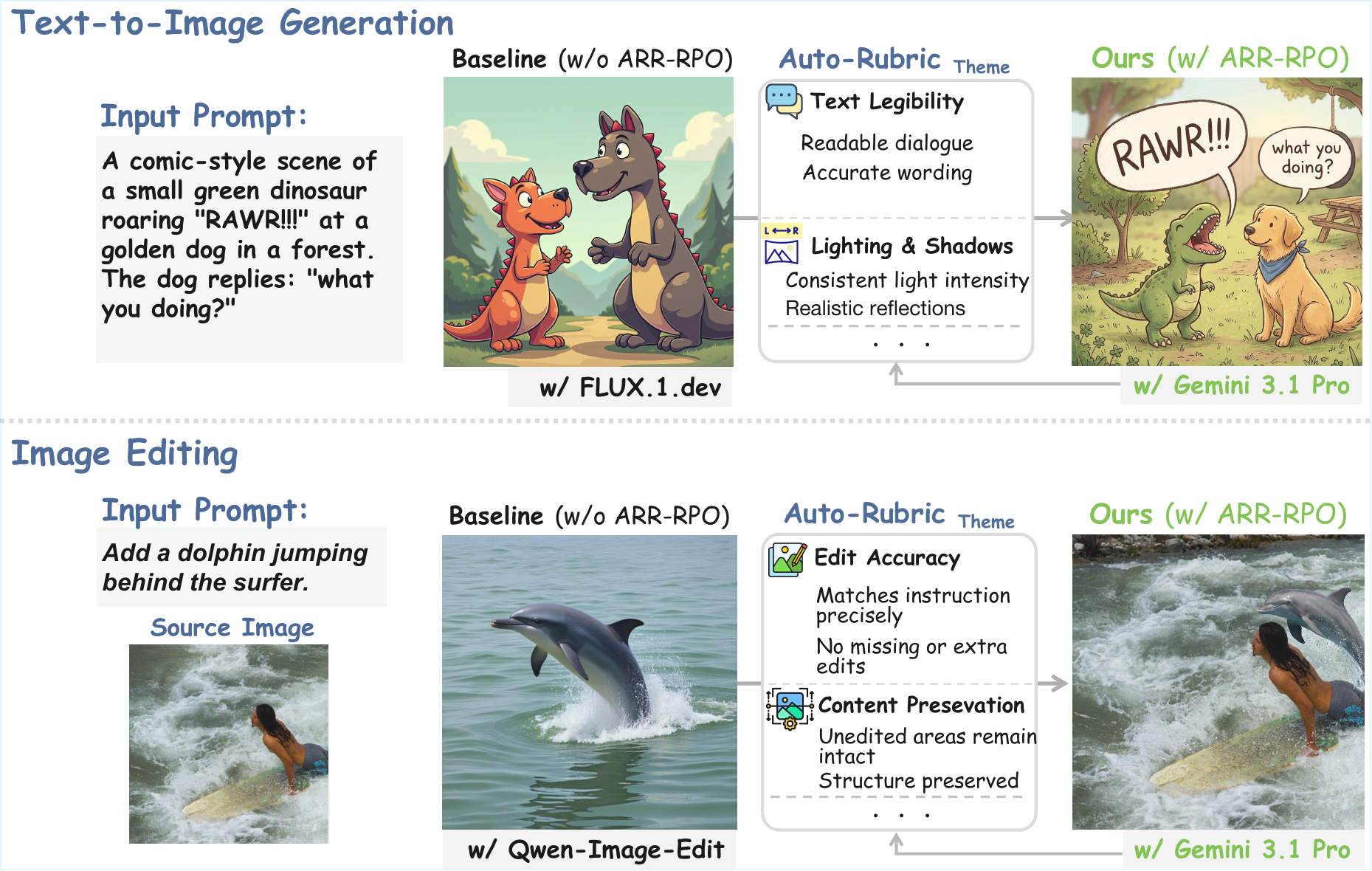}
            \caption{\textbf{Text-to-image and image editing examples} (ARR-RPO Gemini 3.1 Pro).}
            \label{fig:result}
        \end{minipage}%
    }
\end{figure}

We evaluate ARR-RPO on six benchmarks: GenEval~\cite{ghosh2023geneval}, DPG-Bench~\cite{hu2024ella}, TIIF~\cite{wei2025tiifbench}, and UniGenBench++~\cite{wang2026unigenbench} for text-to-image generation; GEdit-Bench~\cite{liu2025step1x} and ImgEdit~\cite{ye2025imgedit} for instruction-guided image editing. ARR-RPO fine-tunes FLUX.1.dev~\cite{labs2025flux} and Qwen-Image-Edit-2509~\cite{wu2025qwen-image} using ARR-generated rubrics as binary reward signals. We instantiate ARR with three VLMs, Qwen3-VL-8B~\cite{bai2025qwen3}, GPT-5~\cite{singh2025openai}, and Gemini~3.1~Pro~\cite{deepmind2026gemini31pro}, to examine how rubric quality scales with judge capability.

\noindent\textbf{Results.}
Figure~\ref{fig:performance} and Table~\ref{tab:generative_main} report generative performance. Two patterns emerge.
First, \textsc{ARR-RPO} consistently outperforms specialist baselines.
For T2I, optimizing FLUX.1.dev with ARR rubrics lifts GenEval (0.66$\rightarrow$0.80), DPG-Bench (83.84$\rightarrow$85.76), TIIF (71.09$\rightarrow$76.85), and UniGenBench++ Short (60.97$\rightarrow$65.89).
In editing, ARR-RPO elevates Qwen-Image-Edit-2509 on GEdit-Bench (7.54$\rightarrow$7.85) and ImgEdit (4.35$\rightarrow$4.43). Second, generated samples (Figure~\ref{fig:result}) exhibit marked improvements in visual quality and edit fidelity, aligning more closely with the multidimensional nature of human preferences.
See Appendix~\ref{tab:generative_app} for full results.
\begin{takeawaybox}
\noindent\textbf{Takeaway:} Rubrics externalize preference structure in a way that makes evaluation directly usable as reward. By grounding policy gradients in rubric-conditioned decisions rather than scalar proxies, ARR-RPO replaces opaque optimization with criteria-aligned improvement, enabling a direct transfer of evaluator fidelity into generative quality without reward model training or architectural change.
%Modeling rubrics as rewards turns improved preference understanding into a direct training signal for generation. By grounding optimization in rubric-conditioned comparisons, ARR-RPO replaces opaque scalar proxies with minimal, criteria-aligned updates, translating evaluator fidelity into consistent gains in generative quality without reward model training or architectural changes.
%Generative alignment is limited less by model capacity than by reward structure. Decomposing preference into explicit criterion-level signals reduces objective interference, and improves compositional fidelity. Rubric fidelity provides a direct pathway from evaluator capability to generation quality, allowing alignment to scale with stronger models and transfer across modalities without architectural changes.
\end{takeawaybox}
\subsection{Ablation analysis}

\begin{figure}[htbp]
    \centering
    \begin{subfigure}[b]{0.48\linewidth}
        \includegraphics[width=\linewidth]{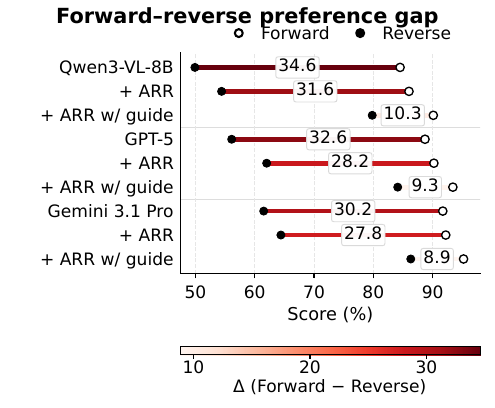}
        \caption{}
        \label{fig:gap}
    \end{subfigure}
    \hfill
    \begin{subfigure}[b]{0.48\linewidth}
        \includegraphics[width=\linewidth]{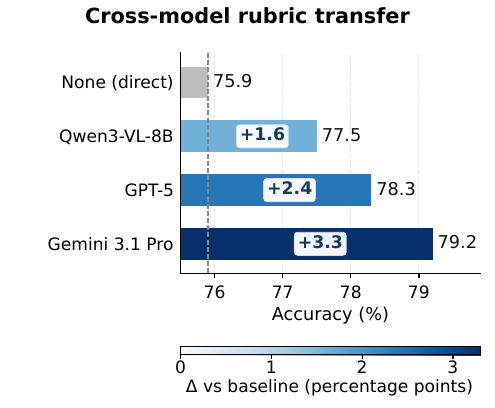}
        \caption{}
        \label{fig:cross_model}
    \end{subfigure}
    \caption{
\textbf{Ablation studies on ARR.}
(a) Forward--reverse preference gaps across evaluators.
(b) Cross-model rubric transfer with a fixed judge.
}\label{fig:combined}
\end{figure}

As shown in Figure~\ref{fig:combined}(a), substantial positional bias ($\Delta = 30.2$ to $34.6$) remains consistent across model scales in the absence of rubric conditioning. This suggests that the bias is not primarily due to insufficient model capacity, but is instead rooted in how preferences are implicitly encoded. Zero-shot ARR provides a modest reduction in bias ($\Delta$ decreases by $3.0$ to $4.8$), while human-guided rubrics lead to a much more pronounced improvement ($\Delta$ reduced to $8.9$ to $10.3$). These results indicate that making evaluation criteria explicit can significantly improve stability.

Figure~\ref{fig:combined}(b) further shows that rubrics generalize across different model families (see Appendix~C). Even when applied to weaker generators, transferred rubrics recover more than half of the performance gap compared to same-family settings. This observation suggests that the effectiveness of ARR is closely related to the quality and structure of the rubric itself, rather than reliance on tight coupling between the generator and the evaluator.

Furthermore, this interpretation is supported by the rubric cardinality ablation (Appendix~\ref{app:cardinality}), where increasing rubric dimensionality consistently improves accuracy, indicating that ARR's gains arise from both finer-grained factorization of preference structure and the quality of the resulting rubric content, rather than model capacity or evaluator--generator coupling.

\section{Conclusion}
\label{conclusion}
We present a unified ARR and RPO framework bridging multimodal preference evaluation and generative alignment. While prevailing approaches rely on implicit, entangled scalar signals that obscure underlying criteria and introduce systematic biases, ARR automatically generates instance-conditioned rubrics by prompting VLMs to externalize latent human preferences into explicit, interpretable criteria. These rubrics provide structured, factorized reward signals for RPO, enabling contrastive preference learning with fine-grained supervision across independent quality dimensions. Together, ARR and RPO replace opaque scalar rewards with explicit, composable criteria, consistently improving both evaluation reliability and generation quality without additional supervision or architectural modifications. This externalization of preference structure offers a principled, scalable pathway toward compositional alignment with nuanced, multidimensional human intent.

%We propose ARR, a framework that leverages automated rubric generation for both evaluating multimodal preferences and aligning generative models. Prevailing methods depend on implicit, entangled preference signals that conceal the underlying criteria and introduce systematic biases. ARR addresses this limitation by prompting VLMs to externalize their latent preference knowledge into explicit, criteria-level rubrics before any comparative judgment is rendered, thereby grounding the evaluation in an interpretable, structured process. These rubrics then define factorized reward signals for generative training, enabling fine-grained preference supervision across independent quality dimensions and promoting compositional alignment with human intent. Across evaluation and generation tasks, this externalization of preference structure consistently enhances reliability and alignment without additional supervision or architectural modification, establishing a principled and scalable mechanism for aligning multimodal systems with the nuanced, multidimensional nature of human preferences.

\FloatBarrier
\begin{ack}
This research was supported in part by Nanyang Technological University through its Nanyang Assistant Professorship Startup Grant (025661-00012). The authors declare no competing interests.
\end{ack}

\clearpage
\bibliographystyle{plain}
\bibliography{reference}
%%%%%%%%%%%%%%%%%%%%%%%%%%%%%%%%%%%%%%%%%%%%%%%%%%%%%%%%%%%%

%%%%%%%%%%%%%%%%%%%%%%%%%%%%%%%%%%%%%%%%%%%%%%%%%%%%%%%%%%%%
%% APPENDIX --- Auto-Rubric as Reward (ARR)
%% NeurIPS 2026 Submission
%%%%%%%%%%%%%%%%%%%%%%%%%%%%%%%%%%%%%%%%%%%%%%%%%%%%%%%%%%%%

\clearpage
\appendix
\section*{Appendix}

\section{Experimental setup details}
\label{app:setup}
This section provides a comprehensive account of the datasets, evaluation protocols, model configurations, training hyperparameters, and computational resources employed throughout the paper. All experiments were conducted on a cluster of 8 NVIDIA H100 (80GB SXM5) GPUs.

\subsection{Datasets}
We evaluate on two families of benchmarks: those designed for assessing preference evaluation fidelity, and those measuring generative quality in text-to-image synthesis and instruction-guided image editing.

\paragraph{Preference evaluation benchmarks.}
\begin{itemize}
    \item \textbf{HPDv3}: A large-scale human preference dataset for text-to-image generation comprising 1.17 million pairwise comparisons collected from diverse user prompts. Each pair presents two images generated from the same prompt, with one image annotated as preferred. We use the official test split and report pairwise preference accuracy.
    \item \textbf{MM-RewardBench2}: A diagnostic benchmark with 4,000 expert-annotated pairwise instances spanning four tasks: text-to-image alignment (T2I), image editing (Edit), visual question answering, and compositional understanding. We report accuracy on the T2I and Edit subtasks separately.
    \item \textbf{EditReward-Bench}: A fine-grained benchmark assessing instruction adherence in image editing, encompassing 13 subtasks with expert human annotations. Each subtask targets a distinct editing operation (e.g., object addition, texture transfer, style modification).
\end{itemize}

\paragraph{Generative benchmarks.}
\begin{itemize}
    \item \textbf{GenEval}: Assesses compositional object accuracy in T2I synthesis by verifying whether generated images contain correct objects and attributes as specified in the prompt. Accuracy is computed via object detection against structured prompt decompositions.
    \item \textbf{DPG-Bench}: Measures alignment with dense, paragraph-length prompts through structured question answering. We report the overall alignment score averaged across all test prompts.
    \item \textbf{TIIF}: Evaluates instruction fidelity across three difficulty tiers (simple, complex, compositional), providing a graded measure of instruction-following capacity. We report the macro-average across tiers.
    \item \textbf{UniGenBench++}: Probes semantic consistency with both Short and Long prompt variants, measuring coherence with brief versus detailed textual descriptions.
    \item \textbf{GEdit-Bench}: A real-world image editing benchmark comprising naturalistic user instructions. Outputs are evaluated by GPT-5 on a 1--10 scale covering instruction adherence, image quality, and preservation of non-targeted regions.
    \item \textbf{ImgEdit}: Evaluates single-turn and multi-turn instruction-driven editing quality using automated metrics and human assessments. We report the composite score averaged across categories and turn depths.
\end{itemize}

\paragraph{RL training datasets}
\begin{itemize}
    \item \textbf{ShareGPT-4o-Image}: A large-scale multimodal corpus for text-to-image generation and editing, containing around 92K high-quality GPT-4o-synthesized samples, including both text-to-image and text-guided image editing pairs, which we use to construct training and evaluation prompts.

\end{itemize}

\subsection{Evaluation protocols}

\paragraph{Preference accuracy.} For all pairwise preference evaluators, we report \emph{preference accuracy}: the proportion of test pairs for which the model assigns a higher reward (or preference) to the human-preferred image. To probe positional robustness, each test pair is evaluated in both its original (forward) and permuted (reverse) order. The gap between forward and reverse accuracy quantifies the degree of position bias.

\paragraph{Generative evaluation.}
For text-to-image generation, FLUX.1-dev uses sampling with 30 sampling steps and a guidance scale of 3.5. For image editing, Qwen-Image-Edit-2509 performs inference with 50 sampling steps and a classifier-free guidance (CFG) scale of 4.0. All benchmark evaluations are conducted using the official evaluation scripts without modification.

\subsection{Model configurations}

\paragraph{ARR instantiation.} Unless otherwise specified, ARR employs a frozen VLM to synthesize five candidate rubrics per labeled training pair. The generation meta-prompt instructs the VLM to decompose the given text prompt into independent evaluative dimensions (e.g., object presence, attribute accuracy, spatial layout, aesthetic quality, instruction adherence), formulating each dimension as a verifiable binary criterion. Rubric synthesis, verification, and refinement are all conducted offline without gradient updates to the judge VLM.

For the ARR~w/ guide variant, the meta-prompt is augmented with a fixed set of human-curated preference exemplars drawn from a held-out subset of the training split of each benchmark. These exemplars consist of (prompt, preferred image, dispreferred image, preference rationale) tuples and are embedded verbatim as in-context demonstrations. No fine-tuning of the VLM is performed; the exemplars serve solely as semantic anchors.

\paragraph{Rubric verification.} Each candidate rubric \(r_i\) generated for a preference pair \((x_i, y_i^+, y_i^-)\) is passed to a separate frozen verifier call, which checks whether applying \(r_i\) as a scoring criterion yields the correct preference decision on the generating pair. If verification fails, we invoke a refinement pass (up to \(T_{\max}=5\) iterations) that presents the verifier's critique alongside the original rubric and requests a revised formulation. Rubrics that remain unverified after \(T_{\max}\) attempts are discarded. In our experiments, approximately \(87\%\) of initial rubrics pass verification without refinement, and fewer than \(4\%\) are ultimately discarded.

\paragraph{Hierarchical structuring.} The verified rubric set \(\mathcal{D}_R\) is organized into a hierarchical prompt structure by a final synthesis call. This call groups criteria by semantic level (coarse: overall alignment; mid: compositional attributes; fine: local details) and orders them by estimated diagnostic value. The resulting structured rubric \(R_{\text{structured}}\) is formatted as a numbered list of axis definitions, each accompanied by a brief operationalization clause. This structure is passed verbatim to the judge VLM as the evaluation conditioning context.

\subsection{Generative training: RPO hyperparameters}

RPO is a reinforcement learning algorithm adapted for denoising diffusion policies in both text-to-image generation (T2I) and image editing. Key hyperparameters are reported in Table~\ref{tab:hparams}.

\begin{table}[htbp]
\centering
\footnotesize
\renewcommand{\arraystretch}{1.15}
\caption{RPO training hyperparameters for T2I (FLUX.1.dev) and image editing (Qwen-Image-Edit-2509).}
\label{tab:hparams}
\begin{tabular}{lcc}
\toprule
\textbf{Hyperparameter} & \textbf{T2I (FLUX.1.dev)} & \textbf{Editing (Qwen-Image-Edit)} \\
\midrule
Base learning rate & $5 \times 10^{-5}$ & $1 \times 10^{-5}$ \\
Batch size & 32 & 16 \\
Candidates per prompt & 2 & 2 \\
Clip $\epsilon$ & 0.2 & 0.2 \\
KL coefficient $\beta$ & 0.01 & 0.02 \\
Positive reward $\lambda$ & 1.0 & 1.0 \\
Negative reward $\gamma$ & 0.1 & 0.1 \\
Denoising steps & 8 & 10 \\
Optimizer & AdamW & AdamW \\
Gradient clipping & 1.0 & 1.0 \\
Trained Parameters & LoRA (rank=16) & LoRA (rank=32) \\
\bottomrule
\end{tabular}
\end{table}

\noindent Training prompts are sampled uniformly from ShareGPT4o-Image, with no data augmentation applied. At each online iteration, two candidate outputs are generated from the current policy $\pi_\theta$, evaluated by the frozen ARR judge conditioned on the structured rubric, and the resulting binary advantage $A \in \{+\lambda, -\gamma\}$ is uniformly distributed across all generation timesteps.
%===========================================================
\section{Auto-Rubric as Reward (ARR) details}
\label{app:arr_details}

This section elaborates on the technical instantiation of Auto-Rubric as Reward (ARR), complementing the concise description provided in Section~\ref{subsec:arr} of the main text. We provide a granular account of the rubric generation pipeline, the verification and refinement protocol, the hierarchical structuring mechanism, and a comparative characterization of ARR within the broader landscape of reward modeling approaches.

\subsection{Rubric generation pipeline}

ARR synthesizes prompt-conditioned rubrics through a three-stage process: \emph{generation}, \emph{verification}, and \emph{structuring}. Each stage is implemented as a frozen (multimodal) large language model call, ensuring that the judge VLM remains unmodified throughout.

\subsubsection{Per-instance rubric generation}
Given a preference pair \((x, y^+, y^-)\) drawn from a pairwise dataset \(\mathcal{D}\), we prompt the generator model \(\mathcal{M}_{\mathrm{gen}}\) to produce a natural language explanation of why \(y^+\) is preferred over \(y^-\). The meta-prompt explicitly instructs the model to:
\begin{itemize}
    \item Decompose the preference into independent, verifiable quality axes (e.g., semantic fidelity, attribute accuracy, spatial coherence).
    \item Formulate each axis as a binary criterion that can be evaluated without reference to the paired candidate.
    \item Avoid holistic or comparative language that presupposes knowledge of both outputs.
\end{itemize}
The resulting rubric \(r_i\) is a structured, axis-wise decomposition of the preference rationale.

\subsubsection{Verification and refinement}
Each candidate rubric \(r_i\) is validated by a separate verifier call \(\mathcal{M}_{\mathrm{verify}}\). The verifier receives the original preference pair \((x, y^+, y^-)\) and the generated rubric \(r_i\), and is tasked with determining whether applying \(r_i\) as an evaluation protocol correctly identifies \(y^+\) as the preferred output. The verification outcome is binary:
\[
v_i = \begin{cases}
\mathrm{true} & \text{if } \mathcal{M}_{\mathrm{verify}}(x, y^+, y^-, r_i) \text{ confirms the original preference}, \\
\mathrm{false} & \text{otherwise}.
\end{cases}
\]
If verification fails, we invoke a refinement pass that supplies the verifier's critique alongside \(r_i\) to a refinement model \(\mathcal{M}_{\mathrm{refine}}\), which produces a revised rubric \(r_i^{(t+1)}\). Refinement iterates up to \(T_{\max}=5\) times; rubrics that remain unverified after this budget are discarded. Empirically, \(87\%\) of initial rubrics pass verification without refinement, and fewer than \(4\%\) are ultimately discarded, attesting to the stability of the generation process.

\subsubsection{Hierarchical structuring}
The verified rubric collection \(\mathcal{D}_R = \{r_i \mid v_i = \mathrm{true}\}\) is subsequently aggregated into a single, hierarchically structured prompt \(R_{\mathrm{structured}}\). The structuring model \(\mathcal{M}_{\mathrm{struct}}\) organizes the rubrics into different evaluation dimensions, including:
\begin{itemize}
    \item \textbf{Overall alignment}: Measures the global consistency between the generated output and the prompt intent.
    \item \textbf{Compositional structure}: Evaluates the presence and relationships of key elements, such as object presence and spatial relations.
    \item \textbf{Fine-grained fidelity}: Focuses on local details and editing-specific accuracy.
    \item \textbf{Other dimensions}: ...
\end{itemize}
The final structured rubric is formatted as a numbered list of evaluation dimensions, where each dimension groups a set of corresponding rubrics. For each dimension, a brief operationalization clause is provided to clarify its focus. This structured rubric is directly used as the conditioning context for the judge VLM during both evaluation and RPO training.

\subsection{Comparative characterization of reward modeling paradigms}

Table~\ref{tab:paradigm_shift} situates ARR within the landscape of contemporary multimodal reward modeling. We contrast ARR against representative pointwise reward models, pairwise reward models, and direct VLM judges. The comparison spans five dimensions: evaluation mode, reward representation, susceptibility to reward hacking, interpretability of the reward signal, and data requirements for training or deployment.

\begin{table}[t]
\centering
\footnotesize
\setlength{\tabcolsep}{3.8pt}
\renewcommand{\arraystretch}{1.2}
\caption{\textbf{Paradigm shift from implicit to explicit reward parameterization.} Comparison of multimodal reward modeling approaches along key operational axes. Pointwise and pairwise reward models require extensive preference data and yield opaque scalar signals. Mix refers to models that support both pairwise and pointwise outputs. ARR uniquely combines zero-shot rubric generation with binary scoring, eliminating training overhead entirely.}
\label{tab:paradigm_shift}
\begin{tabular}{@{}lccccc@{}}
\toprule
\textbf{Method} & \textbf{Mode} & \textbf{Reward Form} & \textbf{Hacking Risk} & \textbf{Interpretability} & \textbf{Data Requirement} \\
\midrule
PickScore & Pointwise & Scalar & High & Low & Large \\
HPSv3 & Pointwise & Scalar & High & Low & Large \\
ImageReward & Pointwise & Scalar & Medium & Low & Large \\
UnifiedReward & Mix & Scalar & Medium & Low & Large \\
VLM-as-Judge & Pairwise (Mix) & Binary & Medium & Medium & Zero-shot \\
\midrule
\textbf{ARR (Ours)} & Pairwise (Mix) & Binary & Low & High & Zero-shot \\
\bottomrule
\end{tabular}
\end{table}

\paragraph{Key distinctions.}
ARR differs from prior approaches in four critical respects:
\begin{enumerate}
    \item \textbf{Zero-shot rubric generation}: ARR synthesizes rubrics on-the-fly from frozen VLMs, enabling immediate deployment in new domains without additional data collection or task-specific supervision.

    \item \textbf{Holistic, rubric-conditioned decision interface}: Rather than aggregating independently scored criteria post hoc, ARR formulates evaluation as a single rubric-conditioned judgment, where all dimensions are jointly considered in a pairwise comparison. This preserves inter-criterion dependencies and avoids inconsistencies introduced by independent scoring and aggregation.

    \item \textbf{Training-free reward interface}: ARR operates without any parameter updates to the judge model, eliminating the computational and data overhead associated with training pointwise or pairwise reward models, while retaining strong generalization through the underlying VLM.

    \item \textbf{Data-efficient rubric induction}: Across all experiments, high-quality rubrics are constructed from as few as 100 preference pairs drawn from ShareGPT-4o-Image. This demonstrates that ARR can recover structured, task-relevant evaluation criteria with minimal supervision, achieving competitive performance with substantially lower data requirements than existing methods.
\end{enumerate}

These properties collectively establish ARR as a lightweight, interpretable, and bias-resilient alternative to both implicit reward models and manually curated rubric systems. Importantly, since ARR builds on a VLM-as-a-judge paradigm, the rubric-conditioned interface is inherently flexible and can be extended beyond pairwise comparison to pointwise scoring or listwise ranking settings. In this work, we focus on the pairwise formulation to isolate and evaluate the robustness of rubric-based decision making under minimal reward hacking risk, providing a controlled setting for studying structured, generative reward modeling in multimodal alignment.

\section{Ablations on position bias in ARR}
\label{app:position_bias}

\subsection{Setup}

Position bias refers to the systematic tendency of a pairwise preference evaluator to favor whichever candidate appears in a fixed ordinal position (e.g., always preferring Image A when presented first), irrespective of actual quality. This constitutes a critical failure mode: an evaluator that achieves high accuracy in the standard presentation order but collapses under permutation produces a spurious reward signal entangled with input ordering rather than genuine quality.

To isolate this effect, we evaluate each pairwise evaluator on the HPDv3 test set under two conditions: (i) \emph{forward order}, where images appear in the original benchmark order; and (ii) \emph{reverse order}, where the two images are swapped. We report forward accuracy (\%), reverse accuracy (\%), and their arithmetic mean (Avg). An ideal, unbiased evaluator would achieve identical accuracy under both conditions. The quantity \(\Delta = \text{Acc}_{\text{fwd}} - \text{Acc}_{\text{rev}}\) serves as our primary measure of positional instability; larger \(\Delta\) indicates stronger position bias.

\subsection{Results}

Table~\ref{tab:position_bias} reports the position bias ablation across three base VLMs and their ARR-augmented variants. All experiments are conducted on the HPDv3 test set.

\begin{table}[htbp]
\centering
\small
\setlength{\tabcolsep}{7pt}
\renewcommand{\arraystretch}{1.18}
\caption{\textbf{Position bias ablation on HPDv3.} Forward and reverse accuracy (\%) are measured by swapping the order of the two images in each preference pair. \(\Delta = \text{Fwd} - \text{Rev}\) quantifies positional instability. ARR variants reduce \(\Delta\) consistently; ARR w/ guide provides the strongest stabilization. Rows are grouped by base model.}
\label{tab:position_bias}
\begin{tabular}{lcccc}
\toprule
\textbf{Method} & \textbf{Forward} & \textbf{Reverse} & \textbf{Avg} & \(\boldsymbol{\Delta}\) \\
\midrule
Qwen3-VL-8B & 84.5 & 49.9 & 67.2 & \cellcolor{red!90}\textcolor{white}{34.6} \\
\quad + ARR & 86.0 & 54.4 & 70.2 & \cellcolor{red!82}\textcolor{white}{31.6} \\
\quad + ARR w/ guide & \textbf{90.1} & \textbf{79.8} & \textbf{85.0} & \cellcolor{red!24}{10.3} \\
\midrule
GPT-5 & 88.7 & 56.1 & 72.4 & \cellcolor{red!85}\textcolor{white}{32.6} \\
\quad + ARR & 90.2 & 62.0 & 76.1 & \cellcolor{red!73}\textcolor{white}{28.2} \\
\quad + ARR w/ guide & \textbf{93.4} & \textbf{84.1} & \textbf{88.8} & \cellcolor{red!21}{9.3} \\
\midrule
Gemini 3.1 Pro & 91.7 & 61.5 & 76.6 & \cellcolor{red!78}\textcolor{white}{30.2} \\
\quad + ARR & 92.2 & 64.4 & 78.3 & \cellcolor{red!72}\textcolor{white}{27.8} \\
\quad + ARR w/ guide & \textbf{95.2} & \textbf{86.3} & \textbf{90.8} & \cellcolor{red!20}{8.9} \\
\bottomrule
\end{tabular}
\end{table}
\subsection{Analysis}

Table~\ref{tab:position_bias} reveals four consistent patterns:

\paragraph{Base VLMs exhibit severe and scale-invariant position bias.} Across all three base models, the gap between forward and reverse accuracy is extreme: \(\Delta = 34.6\) for Qwen3-VL-8B, \(32.6\) for GPT-5, and \(30.2\) for Gemini 3.1 Pro. Crucially, this gap does not diminish with model capability: the most capable model (Gemini 3.1 Pro) yields a marginally smaller but still operationally severe \(\Delta\) of 30.2. This confirms that positional instability is a structural deficiency tied to the implicit parameterization of preference knowledge, not a capacity limitation that resolves with scale.

\paragraph{Zero-shot ARR yields consistent but moderate debiasing.} Conditioning the VLM on auto-generated rubrics reduces \(\Delta\) by 3.0--4.8 points across all three models (e.g., \(34.6 \rightarrow 31.6\) for Qwen3-VL-8B). The mechanism is interpretable: by requiring the model to commit to explicit evaluation criteria before inspecting the candidates, ARR partially anchors the judgment in criterion-level evidence rather than holistic gestalt impressions susceptible to ordering heuristics. However, a substantial gap persists, indicating that self-generated rubrics alone do not fully overcome the structural mismatch between latent preference encoding and stable pairwise judgment.

\paragraph{Preference-conditioned ARR provides qualitatively stronger stabilization.} \texttt{ARR w/ guide} reduces \(\Delta\) dramatically, to \(10.3\), \(9.3\), and \(8.9\) for the three base models respectively, corresponding to reductions of \(24.3\), \(23.3\), and \(21.3\) points relative to the unaugmented baseline. The effect on reverse accuracy is particularly striking: Qwen3-VL-8B improves from \(49.9\%\) (near-random on reversed pairs) to \(79.8\%\), indicating that human preference exemplars substantially enhance the model's capacity to identify quality differences in an order-agnostic manner. This suggests that the key failure mode in unaugmented VLM judges is not an inability to perceive relevant features, but rather an inability to stably weight them independently of presentation order, a failure that explicit, human-grounded rubrics can partially correct.

\paragraph{Residual bias remains non-trivial.} Even under \texttt{ARR w/ guide} with Gemini 3.1 Pro (\(\Delta = 8.9\)), meaningful positional instability persists. A perfectly unbiased evaluator would achieve \(\Delta = 0\). This residual gap underscores that current VLMs do not yet fully ground preference evaluation in stable criteria, and that stronger human preference guidance amplifies the effect of ARR rather than eliminating the need for it. 

\subsection{Cross-model rubric transfer}
\begin{table}[H]
\centering
\caption{Cross-model rubric transfer on HPDv3. The judge is fixed to Gemini~3.1~Pro; only the rubric generator varies. The direct baseline uses no rubric. Accuracy (\%) denotes agreement with human preference labels.}
\label{tab:cross_model}
\begin{tabular}{lcc}
\toprule
\textbf{Rubric Generator} & \textbf{Judge} & \textbf{Accuracy (\%)} \\
\midrule
None (direct) & Gemini~3.1~Pro & 75.9 \\
Qwen3-VL-8B      & Gemini~3.1~Pro & 77.5 \\
GPT-5            & Gemini~3.1~Pro & 78.3 \\
Gemini~3.1~Pro     & Gemini~3.1~Pro & 79.2 \\
\bottomrule
\end{tabular}
\vspace{4pt}
\end{table}

\noindent\textbf{Cross-model rubric transfer.}
To further verify that ARR does not rely on same-family co-adaptation, we fix Gemini~3.1~Pro as the judge and generate rubrics using Qwen3-VL-8B, GPT-5, and Gemini~3.1~Pro.
Table~\ref{tab:cross_model} reports HPDv3 accuracy from a transfer experiment conducted separately from the evaluations in Tables~\ref{tab:evaluator_main} and~\ref{tab:position_bias}.
Even rubrics from the weakest generator, Qwen3-VL-8B, improve accuracy from 75.9\% (direct) to 77.5\%, closing more than half of the gap to same-family rubrics (79.2\%).
This demonstrates that the rubric structure itself, rather than shared model biases, is the primary contributor to evaluation robustness.
%===========================================================
\section{Ablations on rubric cardinality}
\label{app:cardinality}

\subsection{Setup}

The number of candidate rubrics generated offline per labeled training pair represents a key design choice in ARR. Too few rubrics may underspecify the relevant evaluation space, failing to capture important axes of quality; too many may introduce redundant, conflicting, or noisy criteria that degrade the signal-to-noise ratio of the resulting reward. To systematically study this trade-off, we vary the number of rubrics generated per item (\(K \in \{1, 5, 10, 20\}\)) while applying the same hierarchical structuring step to all settings, and measure preference accuracy on the HPDv3 test set. All experiments employ Qwen3-VL-8B-Instruct as the base judge and utilize zero-shot rubric generation (i.e., without human preference exemplars), thereby isolating the effect of cardinality from that of guidance quality.

\subsection{Results}

Table~\ref{tab:cardinality} reports preference accuracy as a function of rubric cardinality. Accuracy is reported as the average of forward and reverse evaluation conditions to ensure that gains are not confounded by positional bias.

\begin{table}[htbp]
\centering
\small
\setlength{\tabcolsep}{9pt}
\renewcommand{\arraystretch}{1.18}
\caption{\textbf{Rubric cardinality vs.\ preference accuracy (HPDv3, Qwen3-VL-8B-Instruct, ARR zero-shot).} Accuracy (\%) reported as average of forward and reverse conditions. \(K=5\) is used as the default in all main experiments.}
\label{tab:cardinality}
\begin{tabular}{cc}
\toprule
\textbf{Num.\ Rubrics per Item (\(K\))} & \textbf{HPDv3 Accuracy (\%)} \\
\midrule
1 & 69.8 \\
5 & 70.2 \\
10 & 72.1 \\
20 & 74.4 \\
\bottomrule
\end{tabular}
\end{table}

\subsection{Analysis}

\paragraph{Accuracy improves monotonically with rubric cardinality.} Increasing \(K\) from \(1\) to \(20\) yields a consistent improvement from \(69.8\%\) to \(74.4\%\), a net gain of \(4.6\) percentage points. This monotonic trend indicates that additional rubric dimensions provide genuinely complementary information rather than merely redundant coverage: each additional axis captures aspects of quality that are not fully addressed by fewer criteria, leading to more discriminative and robust evaluations.

\paragraph{The gains from \(K = 1\) to \(K = 5\) are modest but non-negligible.} The increment from a single rubric to five rubrics yields only \(0.4\) percentage points, suggesting that a well-formed single rubric already captures the primary quality axis relevant to a given preference pair. However, the subsequent gains from \(K = 5\) to \(K = 10\) (\(+1.9\) points) and from \(K = 10\) to \(K = 20\) (\(+2.3\) points) demonstrate that finer-grained decomposition becomes increasingly consequential for difficult pairs where the quality differential is subtle or multidimensional.

\paragraph{Practical trade-offs and the choice of \(K = 5\).} A larger \(K\) increases offline generation and verification cost. Each ordered pair presentation uses one judge call with the cached task rubric. We find that \(K = 5\) provides a favorable accuracy--efficiency trade-off, achieving \(70.2\%\) accuracy with modest offline computational overhead. For deployment contexts where the offline rubric construction budget is constrained, \(K = 5\) constitutes a well-calibrated default. For high-stakes evaluation scenarios, \(K = 20\) delivers the strongest performance in this ablation.

\paragraph{Noise considerations at high cardinality.} We note that despite the accuracy gains, larger \(K\) also elevates the probability of including noisy or redundant criteria: the marginal rubric at \(K = 20\) is necessarily less discriminative than the most salient rubric at \(K = 1\). In RPO training, noisy rubric axes contribute low-magnitude gradient signals whose impact is diluted through averaging across axes during reward aggregation, without harming overall convergence.
%===========================================================
\section{Rubric Policy Optimization details}
\label{app:rpo_details}

\subsection{Algorithm overview}
\begin{algorithm}[t]
\caption{Rubric Policy Optimization (RPO)}
\label{alg:rpo}
\begin{algorithmic}[1]
\REQUIRE Pretrained policy \(\pi_{\theta_0}\), reference policy \(\pi_{\mathrm{ref}}\), frozen ARR judge \(\mathcal{M}_{\theta}\), training prompt distribution \(\mathcal{D}\), number of iterations \(N\), batch size \(B\), positive reward magnitude \(\lambda\), negative reward magnitude \(\gamma\), Clip threshold \(\epsilon\), KL coefficient \(\beta\)
\ENSURE Optimized policy \(\pi_{\theta_N}\)
\FOR{iteration \(k = 1, \dots, N\)}
    \STATE Sample a batch of prompts \(\{h_j\}_{j=1}^B\) from \(\mathcal{D}\)
    \FOR{each prompt \(h_j\)}
        \STATE Generate two candidate outputs: \(y_j^1, y_j^2 \sim \pi_{\theta_{k-1}}(\cdot \mid h_j)\)
        \STATE Set \(R_j \leftarrow R_{\mathrm{structured}}\), where \(R_{\mathrm{structured}}\) is the fixed cached task rubric \COMMENT{via ARR (Section~\ref{subsec:arr})}
        \STATE Obtain binary preference: \(p_j = \mathcal{M}_{\theta}(y_j^1 \succ y_j^2 \mid h_j, R_j)\)
        \STATE Assign advantages: \(A_j^{\mathrm{win}} \leftarrow +\lambda\), \(A_j^{\mathrm{loss}} \leftarrow -\gamma\)
        \STATE Distribute advantage uniformly across all generation timesteps \(t = 0, \dots, T-1\)
    \ENDFOR
    \STATE Compute PPO-clipped objective \(\mathcal{L}_{\mathrm{RPO}}(\theta_{k-1})\) according to Equation~(\ref{eq:rpo_objective})
    \STATE Update policy: \(\theta_k \leftarrow \theta_{k-1} + \eta \nabla_\theta \mathcal{L}_{\mathrm{RPO}}(\theta_{k-1})\)
\ENDFOR
\RETURN \(\pi_{\theta_N}\)
\end{algorithmic}
\end{algorithm}
RPO is an online policy gradient algorithm that leverages ARR-generated rubrics as binary reward signals to align a generative policy \(\pi_{\theta}\) with multidimensional human preferences. As described in Section~\ref{subsec:rpo} of the main text, RPO operates in a fully online fashion: at each iteration, it (i) samples prompts from a training distribution \(\mathcal{D}\), (ii) generates two candidate outputs from the current policy, (iii) evaluates the candidates using the frozen ARR judge conditioned on the cached task rubric, and (iv) updates \(\pi_{\theta}\) via a PPO-style policy gradient with KL regularization. The full training procedure is formalized in Algorithm~\ref{alg:rpo}.

\subsection{KL regularization and training stability}

The KL penalty \(\beta D_{\mathrm{KL}}(\pi_{\theta} \| \pi_{\mathrm{ref}})\) in the RPO objective (Equation~\ref{eq:rpo_objective} in the main text) serves two purposes. First, it prevents excessive policy drift away from the pretrained reference distribution \(\pi_{\mathrm{ref}}\), preserving the generative priors learned during pretraining. Second, it stabilizes training by bounding the entropy reduction induced by reward maximization, thereby mitigating mode collapse. We set \(\beta = 0.01\) for T2I and \(\beta = 0.02\) for image editing; the higher editing coefficient reflects the narrower action space and greater susceptibility to distributional collapse.

We observe that RPO training exhibits substantially lower variance in reward trajectories compared to reward-model-based RL baselines. We attribute this stability to two factors: (i) the frozen nature of the ARR judge eliminates reward model drift as a source of instability, and (ii) the rubric-conditioned binary signal provides a more consistent gradient direction than scalar reward models, which collapse multi-dimensional quality into a single value subject to distributional shift.
\section{Limitation}
\paragraph{Frozen-model focus and fine-tuning potential.} 
This work deliberately concentrates on frozen multimodal foundation models to isolate the effect of externalizing latent preference knowledge through auto-generated rubrics, rather than through parameter updates. By converting implicit, entangled preference representations into explicit, independently verifiable criteria, these rubrics serve as a structured interface that systematically suppresses evaluation biases, enhances interpretability, and directly translates into more reliable and hack-resistant reward signals for generative alignment. The demonstration that this training-free rubric conditioning alone can outperform dedicated pairwise reward models underscores the overlooked significance of the preference interface itself: the primary bottleneck in multimodal alignment is not a deficit of model capacity, but the absence of a stable, factorized criterion space for applying it. While further fine-tuning of the underlying VLMs would likely improve rubric fidelity and downstream generative quality, our results establish that the core value of the rubric paradigm lies in making preference evaluation structurally transparent, bias-resilient, and scalable across tasks without requiring any modification to the judge model.

\paragraph{Pairwise formulation as a robustness measure.} 
We adopt pairwise comparison as the core evaluation protocol because its comparative nature offers stronger structural resistance to reward hacking than pointwise scoring or differentiable reward models. Conditioning these judgments on explicit, prompt specific rubrics amplifies this resilience by grounding evaluation in inspectable, independently verifiable criteria that leave little room for opaque manipulation. This rubric driven interface also endows ARR with considerable extensibility: because the criteria are expressed in natural language, they can be dynamically expanded, refined, or adapted to new domains without any retraining of the underlying judge. The rubric thus functions as a transparent, stable scaffold that decouples evaluation logic from model parameters, preserving interpretability even as task requirements evolve. Although generative reward models and end to end VLM judges offer alternative paradigms, we prioritize the pairwise rubric interface as the most defensible, interpretable, and hack resistant configuration within the current alignment landscape, precisely because it externalizes the preference structure that other approaches leave implicit.

\paragraph{Human supervision and self-improvement.} 
While the ARR framework readily accommodates human supervision to further refine rubric quality and specificity, the present work deliberately emphasizes what can be achieved with no additional annotation. Our experiments demonstrate that multimodal foundation models can substantially self-improve their comprehension and reasoning over human preferences purely through the auto-rubric process, using only self-generated criteria to guide evaluation and optimization. This finding is significant because it reveals that the rubric mechanism itself, even without curated exemplars, provides a sufficient and scalable structure for preference alignment, transforming latent knowledge into actionable, verifiable constraints. Nevertheless, we acknowledge that fully automated rubric generation may not yet reach the precision or domain-specific nuance that curated human guidance could provide, and we therefore treat the deeper integration of human-in-the-loop rubric curation as a natural and valuable extension. The present results establish a lower bound: even in the absence of human intervention, externalizing preference structure through auto-rubrics proves remarkably effective, while the upper bound accessible through human refinement remains an open and promising direction.

\clearpage
\section{Image generation and editing examples}
\begin{figure}[H]
    \centering
    \makebox[\textwidth][c]{%
        \begin{minipage}{1\textwidth}
            \centering
            \includegraphics[width=\linewidth]{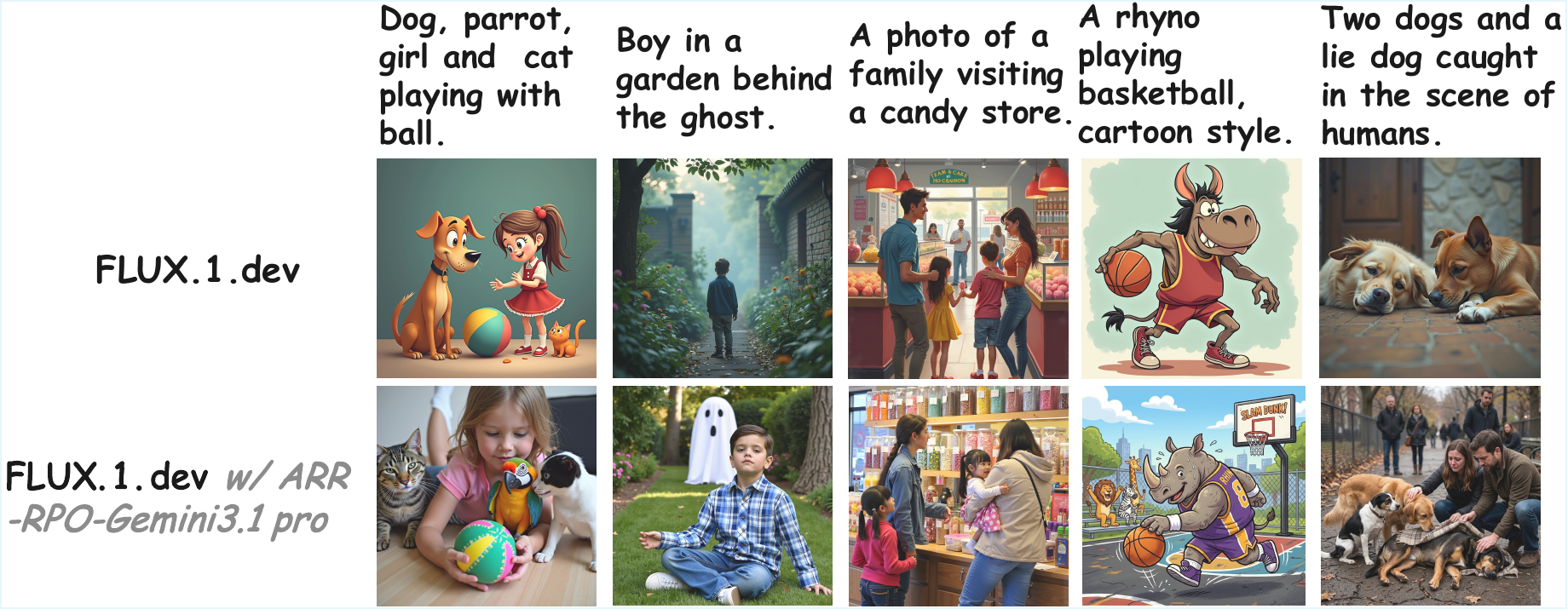}
            \caption{\textbf{Examples of text-to-image generation.}}
            \label{fig:t2i_ex}
        \end{minipage}%
    }
\end{figure}

\begin{figure}[H]
    \centering
    \makebox[\textwidth][c]{%
        \begin{minipage}{1\textwidth}
            \centering
            \includegraphics[width=\linewidth]{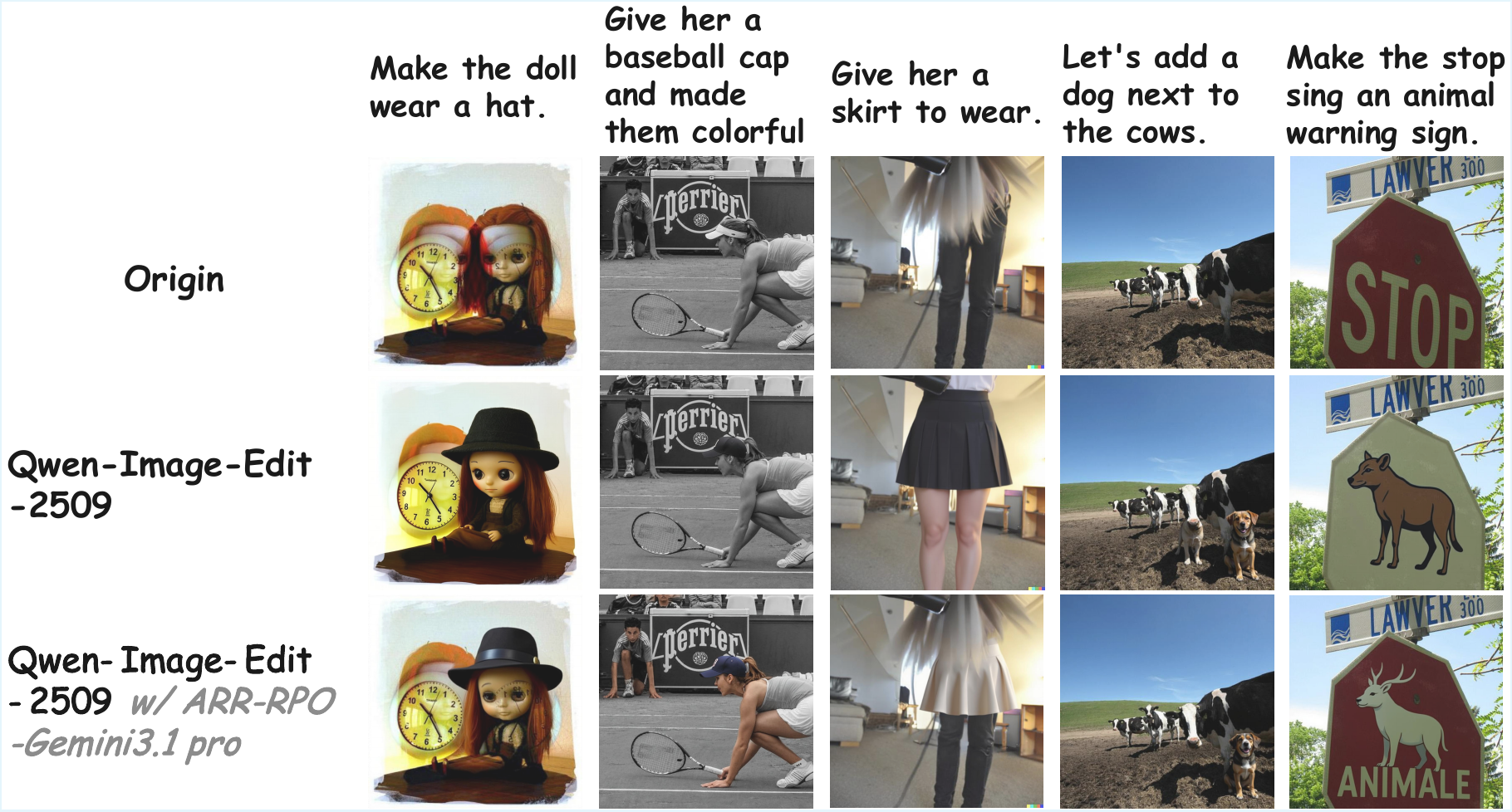}
            \caption{\textbf{Examples of image editing.}}
            \label{fig:edit_ex}
        \end{minipage}%
    }
\end{figure}

\newpage
\section{Full results}
\subsection{Image generation and editing}
\begin{table}[H]
\centering
\footnotesize
\setlength{\tabcolsep}{3.6pt}
\renewcommand{\arraystretch}{0.95}
\caption{\textbf{Generative performance across T2I and image editing benchmarks.} Blue-shaded rows denote ARR-RPO. Green arrows indicate absolute gains over the baseline.}
\label{tab:generative_app}
\begin{adjustbox}{max width=\textwidth}
\begin{tabular}{lccccccc}
\toprule
\multirow{2}{*}{Method} & \multicolumn{5}{c}{Text-to-Image} & \multicolumn{2}{c}{Image Editing} \\
\cmidrule(lr){2-6} \cmidrule(lr){7-8}
& GenEval & DPG-Bench & TIIF & \multicolumn{2}{c}{UniGenBench++} & GEdit-Bench & ImgEdit \\
\cmidrule(lr){5-6}
& & & & Short & Long & & \\
\midrule
\multicolumn{8}{l}{\textbf{\textit{Specialist Model (T2I)}}} \\
SDXL\cite{podell2023sdxl} & 0.55 & 74.65 & 54.96 & 40.22 & 41.48 & --- & --- \\
Emu3\cite{wang2024emu3} & 0.54 & 80.60 & -- & 45.42 & 50.59 & --- & --- \\
JanusFlow\cite{ma2025janusflow} & 0.63 & 79.68 & -- & 47.10 & 54.80 & --- & --- \\
FLUX.1-Dev & 0.66 & 83.84 & 71.09 & 60.97 & 69.42 & --- & --- \\
DALLE-3\cite{betker2023dalle3} & 0.67 & 83.50 & 74.96 & 68.85 & 70.82 & --- & --- \\
BLIP3o-4B\cite{chen2025blip3} & 0.81 & 79.36 & -- & 59.57 & 61.01 & --- & --- \\
Janus-Pro-7B\cite{chen2025januspro} & 0.80 & 84.19 & 66.50 & 61.36 & 71.11 & --- & --- \\
Show-o2\cite{xie2025showo2} & 0.76 & 86.14 & -- & 61.90 & 70.33 & --- & --- \\
OmniGen2\cite{wu2025omnigen2} & 0.80 & 83.57 & -- & 63.09 & 71.39 & --- & --- \\
BAGEL\cite{deng2025bagel} & 0.82 & 85.07 & 71.50 & 59.91 & 71.26 & --- & --- \\
\midrule
\rowcolor{cyan!5}\multicolumn{8}{l}{\textbf{\textit{ARR-RPO / T2I (Ours)}}} \\
\rowcolor{cyan!5} w/ RPO-Qwen3vl-8B      & \gain{0.72}{0.06} & \gain{84.67}{0.83} & \gain{73.81}{2.72} & \gain{63.28}{2.31} & \gain{71.05}{1.63} & --- & --- \\
\rowcolor{cyan!5} w/ RPO-Qwen3vl-8B-ARR  & \gain{0.74}{0.08} & \gain{85.03}{1.19} & \gain{74.92}{3.83} & \gain{64.17}{3.20} & \gain{71.82}{2.40} & --- & --- \\
\rowcolor{cyan!5} w/ RPO-GPT-5           & \gain{0.76}{0.10} & \gain{84.97}{1.13} & \gain{74.84}{3.75} & \gain{64.22}{3.25} & \gain{71.78}{2.36} & --- & --- \\
\rowcolor{cyan!5} w/ RPO-GPT-5-ARR       & \gain{0.78}{0.12} & \gain{85.41}{1.57} & \gain{76.18}{5.09} & \gain{65.36}{4.39} & \gain{72.41}{2.99} & --- & --- \\
\rowcolor{cyan!5} w/ RPO-Gemini 3.1 Pro      & \gain{0.77}{0.11} & \gain{85.02}{1.18} & \gain{75.69}{4.60} & \gain{64.76}{3.79} & \gain{72.13}{2.71} & --- & --- \\
\rowcolor{cyan!5} w/ RPO-Gemini 3.1 Pro-ARR  & \gain{0.80}{0.14} & \gain{85.76}{1.92} & \gain{76.85}{5.76} & \gain{65.89}{4.92} & \gain{72.93}{3.51} & --- & --- \\
\midrule
\multicolumn{8}{l}{\textbf{\textit{Specialist Model (Editing)}}} \\
Instruct-Pix2Pix\cite{brooks2023instructpix2pix} & --- & --- & --- & --- & --- & 3.68 & 1.88 \\
AnyEdit\cite{yu2025anyedit} & --- & --- & --- & --- & --- & 3.21 & 2.45 \\
Step1X-Edit\cite{liu2025step1x} & --- & --- & --- & --- & --- & 6.97 & 3.06 \\
Qwen-Image-Edit-2509\cite{wu2025qwen-image} & --- & --- & --- & --- & --- & 7.54 & 4.35 \\
UniWorld-Qwen-Image-Edit\cite{li2025uniworldv2} & --- & --- & --- & --- & --- & 7.76 & 4.48 \\
\midrule
\rowcolor{cyan!5}\multicolumn{8}{l}{\textbf{\textit{ARR-RPO / Image Editing (Ours)}}} \\
\rowcolor{cyan!5} w/ RPO-Qwen3vl-8B      & --- & --- & --- & --- & --- & \gain{7.63}{0.09} & \gain{4.37}{0.02} \\
\rowcolor{cyan!5} w/ RPO-Qwen3vl-8B-ARR  & --- & --- & --- & --- & --- & \gain{7.66}{0.12} & \gain{4.38}{0.03} \\
\rowcolor{cyan!5} w/ RPO-GPT-5           & --- & --- & --- & --- & --- & \gain{7.65}{0.11} & \gain{4.38}{0.03} \\
\rowcolor{cyan!5} w/ RPO-GPT-5-ARR       & --- & --- & --- & --- & --- & \gain{7.72}{0.18} & \gain{4.40}{0.05} \\
\rowcolor{cyan!5} w/ RPO-Gemini 3.1 Pro      & --- & --- & --- & --- & --- & \gain{7.79}{0.25} & \gain{4.39}{0.04} \\
\rowcolor{cyan!5} w/ RPO-Gemini 3.1 Pro-ARR  & --- & --- & --- & --- & --- & \gain{7.85}{0.31} & \gain{4.43}{0.08} \\
\bottomrule
\end{tabular}
\end{adjustbox}
\end{table}
\clearpage
\vspace{0.75em}

\begin{table}[H]
\centering
\footnotesize
\setlength{\tabcolsep}{4.2pt}
\renewcommand{\arraystretch}{1.5}
\caption{\textbf{Additional experimental results: Post-training performance of BAGEL using ARR-RPO across T2I benchmarks.} Best in \textbf{bold}. Blue-shaded rows denote ARR-RPO variants. Green arrows indicate absolute gains over the BAGEL baseline.}
\label{tab:additional_bagel}
\begin{tabular}{lccccc}
\toprule
Method & GenEval & DPG-Bench & TIIF & \multicolumn{2}{c}{UniGenBench++} \\
\cmidrule(lr){5-6}
& & & & Short & Long \\
\midrule
\multicolumn{6}{l}{\textbf{\textit{BAGEL (Baseline)}}} \\
BAGEL\cite{deng2025bagel} & 0.82 & 85.07 & 71.50 & 59.91 & 71.26 \\
\midrule
\rowcolor{cyan!5}\multicolumn{6}{l}{\textbf{\textit{ARR-RPO / T2I (Ours)}}} \\
\rowcolor{cyan!5}w/ RPO-Qwen3vl-8B      & \gain{0.85}{0.03} & \gain{85.45}{0.38} & \gain{73.12}{1.62} & \gain{62.34}{2.43} & \gain{71.89}{0.63} \\
\rowcolor{cyan!5}w/ RPO-Qwen3vl-8B-ARR  & \gain{0.88}{0.06} & \gain{85.82}{0.75} & \gain{74.05}{2.55} & \gain{63.05}{3.14} & \gain{72.35}{1.09} \\
\rowcolor{cyan!5}w/ RPO-GPT-5           & \gain{0.86}{0.04} & \gain{85.91}{0.84} & \gain{74.48}{2.98} & \gain{63.72}{3.81} & \gain{72.58}{1.32} \\
\rowcolor{cyan!5}w/ RPO-GPT-5-ARR       & \gain{0.90}{0.08} & \gain{86.28}{1.21} & \gain{75.62}{4.12} & \gain{64.81}{4.90} & \gain{73.15}{1.89} \\
\rowcolor{cyan!5}w/ RPO-Gemini 3.1 Pro      & \gain{0.87}{0.05} & \gain{86.15}{1.08} & \gain{75.41}{3.91} & \gain{64.62}{4.71} & \gain{72.97}{1.71} \\
\rowcolor{cyan!5}w/ RPO-Gemini 3.1 Pro-ARR  & \gain{\textbf{0.92}}{0.10} & \gain{\textbf{86.74}}{1.67} & \gain{\textbf{76.85}}{5.35} & \gain{\textbf{65.92}}{6.01} & \gain{\textbf{73.82}}{2.56} \\
\bottomrule
\end{tabular}
\end{table}

\subsection{Human preference}
\begin{table}[H]
\centering
\footnotesize
\setlength{\tabcolsep}{4.5pt}
\renewcommand{\arraystretch}{0.9}
\caption{\textbf{Evaluator performance across four preference benchmarks.} Accuracy (\%) denotes agreement with human preference labels. The best result in each column is \textbf{bold}. Blue-shaded rows indicate ARR. Green arrows indicate absolute gains over the baseline.}
\label{tab:evaluator_app}
\begin{tabular}{lcccc}
\toprule
\multirow{2}{*}{Method} & HPDv3 & MM-RewardBench2 & MM-RewardBench2 & EditReward-Bench \\
& & (T2I) & (Edit) & \\
& Acc. & Acc. & Acc. & Acc. \\
\midrule
\multicolumn{5}{l}{\textbf{\textit{Trained Reward Model}}} \\
CLIPScore\cite{hessel2022clipscore} & 48.6 & 51.0 & --- & --- \\
PickScore\cite{kirstain2023pick} & 65.6 & 58.6 & --- & --- \\
ImageReward\cite{xu2023imagereward} & 58.6 & 54.0 & --- & --- \\
UnifiedReward\cite{wang2025unified} & 66.0 & 59.8 & --- & --- \\
UnifiedReward-Thinking\cite{wang2025unirewardthinking} & 68.1 & 66.0 & --- & --- \\
HPSv2\cite{wu2023hps2} & 65.3 & 54.7 & --- & --- \\
HPSv3\cite{ma2025hpsv3} & 76.9 & 60.2 & --- & --- \\
EditReward\cite{wu2025editreward} & --- & --- & 67.2 & 56.45 \\
\midrule
\multicolumn{5}{l}{\textbf{\textit{VLM-as-Judge (Direct)}}} \\
Qwen3-VL-8B & 67.2 & 57.6 & 59.2 & 54.01 \\
GPT-5 & 72.4 & 70.5 & 73.8 & 57.53 \\
Gemini~3.1~Pro & 76.6 & 75.1 & 77.4 & 61.23 \\
\midrule
\rowcolor{cyan!5}\multicolumn{5}{l}{\textbf{\textit{ARR (Ours)}}} \\
\rowcolor{cyan!5}Qwen3vl-8B + ARR & \gainstd{70.2}{0.2}{3.0} & \gainstd{62.7}{0.2}{5.1} & \gainstd{65.5}{0.3}{6.3} & \gainstd{57.22}{0.1}{3.21} \\
\rowcolor{cyan!5}GPT-5 + ARR & \gainstd{76.1}{0.2}{3.7} & \gainstd{74.7}{0.4}{4.2} & \gainstd{77.5}{0.3}{3.7} & \gainstd{61.01}{0.1}{3.48} \\
\rowcolor{cyan!5}Gemini 3.1 Pro + ARR & \gainstd{\textbf{78.3}}{0.1}{1.7} & \gainstd{\textbf{78.9}}{0.2}{3.8} & \gainstd{\textbf{79.2}}{0.2}{1.8} & \gainstd{\textbf{63.27}}{0.2}{2.04} \\
\bottomrule
\end{tabular}
\end{table}

\clearpage
\section{Prompts and rubrics}
\label{app:example_rubrics}
\begin{tcolorbox}[breakable,colback=blue!2,colframe=takeawayline,boxrule=0.5pt,
  title={T2I rubrics (Gemini 3.1 Pro)},title after break={T2I rubrics (Gemini 3.1 Pro) (continued)},fonttitle=\bfseries,
  left=6pt,right=6pt,top=6pt,bottom=6pt]
\begin{lstlisting}[basicstyle=\ttfamily\footnotesize,breaklines=true,breakatwhitespace=false,
  columns=fullflexible,keepspaces=true,showstringspaces=false,aboveskip=0pt,belowskip=0pt]
Rubric 1:
Theme: Architectural and Structural Fidelity
- Tip1: Assess clarity and precision of building facades, windows, railings, and structural elements for geometric accuracy and absence of
distortion.
- Tip2: Evaluate consistency of architectural details (e.g., columns, arches, textures) across the scene without illogical or impossible
geometry.
- Tip3: Check for realistic material rendering (wood grain, stone patina, metal sheen) and absence of overly smooth or artificial surfaces.
- Tip4: Ensure environmental context is coherent, with buildings, paths, and landscapes logically integrated and proportionally accurate.
Rubric 2:
Theme: Lighting, Shadow, and Reflection Realism
- Tip1: Evaluate how naturally light interacts with surfaces, including direction, intensity, and shadow gradients that match the scene's
source.
- Tip2: Assess reflections on water, glass, and metallic surfaces for physical plausibility and consistency with ambient lighting.
- Tip3: Check for believable depth and dimensionality through consistent, directional shadows and highlights across all objects and subjects.
- Tip4: Identify flat, unnatural, or inconsistent lighting that flattens the scene or creates artificial glow and highlights.
Rubric 3:
Theme: Texture and Material Fidelity
- Tip1: Judge the clarity and detail of fine textures (fabric weave, fur, wood grain, skin pores) without smearing or over-smoothing.
- Tip2: Assess realism of material properties (metallic sheen, fabric wear, plastic gloss) and absence of uniform, plastic-like surfaces.
- Tip3: Look for natural imperfections (dust, scratches, frayed edges) that enhance authenticity and avoid overly pristine or artificial
finishes.
- Tip4: Evaluate surface variations and non-repetitive patterns to distinguish between realistic textures and AI-generated uniformity.
Rubric 4:
Theme: Anatomical and Object Plausibility
- Tip1: Assess anatomical accuracy of human and animal forms, including proportions, posture, and facial features for naturalism.
- Tip2: Evaluate mechanical and structural plausibility of objects (tools, vehicles, clothing) for logical construction and physical
consistency.
- Tip3: Check for distorted limbs, unnatural hand shapes, or illogical object boundaries that break immersion or suggest AI artifacts.
- Tip4: Ensure subjects interact with their environment in believable ways (e.g., shadows, occlusion, posture) without surreal or impossible
elements.
Rubric 5:
Theme: Visual Coherence and Composition
- Tip1: Assess overall scene balance, depth, and arrangement of elements for aesthetic harmony and intentional design.
- Tip2: Evaluate framing, perspective, and depth of field for natural spatial relationships and absence of awkward or cluttered compositions.
- Tip3: Judge the integration of foreground, midground, and background for visual flow and environmental immersion.
- Tip4: Look for compositional clarity, with focal points and visual hierarchy that guide the viewer's eye without distraction.
Rubric 6:
Theme: AI Artifact Detection and Elimination
- Tip1: Identify telltale signs of AI generation: unnatural edges, repeated patterns, inconsistent perspective, or surreal details.
- Tip2: Look for artifacts such as smearing, halos, pixelation, or warped geometry that disrupt visual authenticity.
- Tip3: Assess for illogical object placement, duplicated elements, or inconsistent lighting that break scene coherence.
- Tip4: Prioritize images with minimal or no AI artifacts, especially in critical areas like anatomy, textures, and object boundaries.
Rubric 7:
Theme: Environmental and Background Realism
- Tip1: Evaluate the realism of natural elements (trees, water, clouds, terrain) for accurate physics, textures, and atmospheric perspective.
- Tip2: Assess background depth and context for logical scale, perspective, and integration with foreground subjects.
- Tip3: Check for believable environmental details (e.g., foliage, ground texture, sky gradients) without AI-style repetition or over-
saturation.
- Tip4: Ensure background elements (buildings, landscapes, props) are contextually plausible and not visually disconnected from the scene.
Rubric 8:
Theme: Text, Typography, and Detail Legibility
- Tip1: Assess clarity and legibility of text, including font consistency, spacing, and absence of smudging or distortion.
- Tip2: Evaluate spelling accuracy, grammar, and semantic coherence of written elements (e.g., signs, captions, product labels).
- Tip3: Check for clean, well-defined typography with appropriate contrast and alignment against background elements.
- Tip4: Prioritize images with crisp, readable text and no garbled, pixelated, or AI-generated text artifacts.
Rubric 9:
Theme: Color, Gradient, and Aesthetic Harmony
- Tip1: Assess color palette richness, saturation, and harmony for contextually appropriate and visually pleasing transitions.
- Tip2: Evaluate natural lighting and color grading for believable gradients, shadows, and reflections without artificial neon or
oversaturation.
- Tip3: Judge overall aesthetic appeal, including balance, contrast, and intentional visual design that enhances mood or narrative.
- Tip4: Look for unified color schemes that support the scene's theme without jarring or mismatched tones.
Rubric 10:
Theme: Detail Fidelity and Micro-Texture Rendering
- Tip1: Assess sharpness and clarity of fine details (hair strands, stitching, buttonholes, tool engravings) without AI-style blurring.
- Tip2: Evaluate preservation of intricate textures (fur, fabric, metal, skin) that convey realism and depth without over-smoothing.
- Tip3: Check for consistent rendering of small elements (e.g., droplets, dust, scratches) that add authenticity without artificial
repetition.
- Tip4: Prioritize images where micro-textures are naturally varied and physically plausible, not uniformly or algorithmically generated.
\end{lstlisting}
\end{tcolorbox}
\begin{minipage}{\linewidth}
\captionsetup{hypcap=false}
\captionof{figure}{\textbf{Auto-generated T2I rubrics (Gemini 3.1 Pro).} Example prompt-conditioned rubrics automatically synthesized by ARR for text-to-image evaluation, spanning dimensions such as architectural fidelity, lighting consistency, texture realism, and AI artifact detection.}
\label{fig:arr_t2i}
\end{minipage}
\clearpage

\begin{tcolorbox}[breakable,colback=blue!2,colframe=takeawayline,boxrule=0.5pt,
  title={T2I evaluation system prompt},title after break={T2I evaluation system prompt (continued)},fonttitle=\bfseries,
  left=6pt,right=6pt,top=6pt,bottom=6pt]
\begin{lstlisting}[basicstyle=\ttfamily\footnotesize,breaklines=true,breakatwhitespace=false,
  columns=fullflexible,keepspaces=true,showstringspaces=false,aboveskip=0pt,belowskip=0pt]
Based on the given evaluation rubrics and the caption, please perform a **pairwise
comparison** between the two input images (Image 1 and Image 2) to determine which one has
higher quality.
{task_description_section}
Evaluation Rubrics:
{rubrics}
**Caption: {sample_content}**
## Task Requirements
- You are evaluating two images: Image 1 and Image 2.
- Using the provided evaluation rubrics, determine which image has better overall quality.
- Assign a rank value to each image:
- rank = 1 -> higher quality (better image)
- rank = 2 -> lower quality (worse image)
- The output format should be: [rank_of_Image1, rank_of_Image2]
- Examples:
- If Image 1 is better than Image 2 -> output: [1, 2]
- If Image 2 is better than Image 1 -> output: [2, 1]
- Important: The two rank values **must be different** --- ties are not allowed.
- Focus solely on the image content and quality per the rubrics; do **not** let image order or
position bias your judgment.
## Output Format
Please strictly follow the JSON format below:
{{
"rank": [rank_of_Image1, rank_of_Image2],
"reason": "Detailed explanation of the reasoning and quality assessment for both images"
}}
Key reminders:
- The `rank` array must contain exactly 2 values, one for each image.
- rank = 1 means higher quality; rank = 2 means lower quality.
\end{lstlisting}
\end{tcolorbox}
\begin{minipage}{\linewidth}
\captionsetup{hypcap=false}
\captionof{figure}{\textbf{T2I evaluation system prompt.} The prompt template used to instruct the VLM judge to perform pairwise comparison for text-to-image generation, including task description, output format requirements, and anti-position-bias reminders.}
\label{fig:t2i_prompt}
\end{minipage}
\clearpage

\begin{tcolorbox}[breakable,colback=blue!2,colframe=takeawayline,boxrule=0.5pt,
  title={Image editing rubrics (Gemini 3.1 Pro)},title after break={Image editing rubrics (Gemini 3.1 Pro) (continued)},fonttitle=\bfseries,
  left=6pt,right=6pt,top=6pt,bottom=6pt]
\begin{lstlisting}[basicstyle=\ttfamily\footnotesize,breaklines=true,breakatwhitespace=false,
  columns=fullflexible,keepspaces=true,showstringspaces=false,aboveskip=0pt,belowskip=0pt]
Rubric 1:
Theme: Fidelity of Built Structures After Editing
- Tip 1: Verify that edited building facades, windows, railings, roofs, and other structural elements preserve precise geometric accuracy, with
straight lines, correct proportions, and no visible distortions such as bending, warping, or duplicated edges.
- Tip 2: Ensure that architectural details, including columns, arches, balconies, decorative patterns, and surface textures, remain internally
consistent and logically constructed after editing, without introducing impossible or unstable structures.
- Tip 3: Check that material properties of architectural components, such as wood grain, stone roughness, brick seams, glass transparency, and
metallic reflections, are faithfully preserved or enhanced without becoming unnaturally smooth, plastic-like, or visually artificial.
- Tip 4: Confirm that edits maintain coherent spatial relationships among buildings, pathways, roads, vegetation, and surrounding environmental
elements, ensuring perspective consistency and physically plausible scene layout.
Rubric 2:
Theme: Physically Plausible Illumination Adjustments
- Tip 1: Ensure that edited lighting direction, intensity, exposure, and color temperature remain physically plausible and align with the
original illumination setup or the newly introduced light source.
- Tip 2: Validate that shadows added, removed, or modified remain globally consistent in direction, softness, scale, and occlusion behavior
across all scene objects and surfaces.
- Tip 3: Check that reflections and specular highlights on water, glass, polished metal, or glossy surfaces are updated consistently to match
the revised lighting conditions and viewing angle.
- Tip 4: Avoid introducing flat lighting, artificial glow, or inconsistent highlights during editing.
Rubric 3:
Theme: Material Integrity During Editing
- Tip 1: Confirm that fine textures (fabric, wood grain, skin, fur) are preserved and not blurred or over-smoothed.
- Tip 2: Ensure that material-specific physical properties, including glossiness, roughness, transparency, translucency, and reflectivity,
remain realistic and visually consistent after editing.
- Tip 3: Introduce or maintain natural imperfections (scratches, dust, wear) where appropriate to avoid artificial perfection.
- Tip 4: Avoid repetitive, tiled, or uniform texture artifacts that may be introduced during the editing process, particularly in large
homogeneous regions.
Rubric 4:
Theme: Realism of Subjects and Objects Post-Edit
- Tip 1: Ensure human and animal anatomy remains proportionally accurate after editing (no distorted limbs or features).
- Tip 2: Validate that edited objects, such as tools, clothing, furniture, vehicles, or accessories, preserve logical structure, correct
functionality, and physical plausibility.
- Tip 3: Detect and correct warped shapes, broken boundaries, or blending artifacts.
- Tip 4: Confirm that subjects interact naturally with their environment (contact, shadows, occlusion).
Rubric 5:
Theme: Cohesion and Aesthetic Integrity
- Tip 1: Ensure edits preserve or improve scene balance and visual hierarchy.
- Tip 2: Validate perspective, framing, and depth relationships after modifications.
- Tip 3: Maintain smooth integration between foreground, midground, and background elements, avoiding abrupt transitions, cutout artifacts, or
mismatched blur levels.
- Tip 4: Avoid clutter, awkward cropping, or disruption of focal points.
Rubric 6:
Theme: Elimination of AI and Editing Artifacts
- Tip 1: Identify and remove common artifacts such as smearing, halos, pixelation, jagged edges, or unnatural blur introduced during editing or
generation.
- Tip 2: Avoid introducing repeated patterns, duplicated elements, or inconsistent geometry.
- Tip 3: Ensure seamless blending between edited and original regions, with smooth transitions in texture, lighting, and color.
- Tip 4: Prioritize clean, polished outputs with minimal visible traces of manipulation, ensuring the image appears naturally captured rather
than artificially altered.
Rubric 7:
Theme: Background and Context Integrity
- Tip 1: Ensure environmental elements (e.g., trees, sky, terrain, buildings) remain physically plausible and visually consistent after editing.
- Tip 2: Maintain consistent scale, depth, and perspective between foreground and background.
- Tip 3: Avoid repetitive, tiled, or overly uniform patterns in natural elements, which may indicate synthetic generation.
- Tip 4: Ensure all background details align contextually with the subject, supporting a coherent scene narrative and avoiding mismatched
environments.
Rubric 8:
Theme: Clarity of Edited Text Elements
- Tip 1: Ensure all text remains clearly legible after editing, with crisp edges and consistent, recognizable font styles.
- Tip 2: Preserve correct spelling, grammar, and semantic meaning, avoiding nonsensical or AI-generated text artifacts.
- Tip 3: Maintain proper alignment, spacing, and contrast to ensure readability against varying backgrounds.
- Tip 4: Avoid distortions such as warping, stretching, or irregular baselines that degrade text quality or professionalism.
Rubric 9:
Theme: Harmonized Color Adjustments
- Tip 1: Ensure color palettes remain balanced and contextually appropriate to the scene, avoiding abrupt or unrealistic shifts.
- Tip 2: Maintain smooth and natural gradients, preventing banding, clipping, or harsh transitions between tones.
- Tip 3: Align color grading with the scene's lighting conditions, time of day, and intended mood or atmosphere.
- Tip 4: Avoid oversaturation, color mismatches, or unnatural hues that reduce realism or visual comfort.
Rubric 10:
Theme: Micro-Detail Integrity in Editing
- Tip 1: Preserve sharpness and fidelity in fine details such as hair strands, fabric stitching, surface engravings, or textures.
- Tip 2: Ensure micro-textures remain natural and are not overly smoothed or artificially sharpened during processing.
- Tip 3: Maintain variation and randomness in small-scale details to avoid repetitive or algorithmically generated patterns.
- Tip 4: Prevent loss of high-frequency information, ensuring that intricate regions retain clarity and do not degrade during editing.
\end{lstlisting}
\end{tcolorbox}
\begin{minipage}{\linewidth}
\captionsetup{hypcap=false}
\captionof{figure}{\textbf{Auto-generated image editing rubrics (Gemini 3.1 Pro).} Example prompt-conditioned rubrics automatically synthesized by ARR for image editing evaluation, covering fidelity preservation, material integrity, lighting consistency, and artifact elimination.}
\label{fig:arr_edit}
\end{minipage}
\clearpage

\begin{tcolorbox}[breakable,colback=blue!2,colframe=takeawayline,boxrule=0.5pt,
  title={Image editing evaluation system prompt},title after break={Image editing evaluation system prompt (continued)},fonttitle=\bfseries,
  left=6pt,right=6pt,top=6pt,bottom=6pt]
\begin{lstlisting}[basicstyle=\ttfamily\footnotesize,breaklines=true,breakatwhitespace=false,
  columns=fullflexible,keepspaces=true,showstringspaces=false,aboveskip=0pt,belowskip=0pt]
Please perform a pairwise comparison between the two edited versions (**Image 1** and **Image 2**)
based on the given rubrics and caption, using the **original image (Image BASE)** as the
reference benchmark. Determine which edited result is superior.
{task_description_section}
**Evaluation Rubrics:**
{rubrics}
**caption: {sample_content}**
## Task Requirements
- You are evaluating two image editing results (Image 1 and Image 2) with **Image BASE (the
original image)** as the ground-truth reference.
- Compare them across the following aspects:
- Fidelity to the original content
- Accuracy and naturalness of instruction following
- Assign rank values to the two edited images:
- rank=1 means better quality (higher fidelity to Image BASE + better editing)
- rank=2 means worse quality
- Output format must be: [rank_of_Image1, rank_of_Image2]
- Examples:
- If Image 1 is better -> [1, 2]
- If Image 2 is better -> [2, 1]
- Important: The two ranks must be different (strictly 1 and 2). All judgments must explicitly
reference Image BASE.
- Judge solely based on image content, the provided rubrics, and comparison with the original
Image BASE. Do not be influenced by image position or order.
## Output Format
Please strictly follow this JSON format:
{
"rank": [rank of Image 1, rank of Image 2],
"reason": "Detailed explanation of the judgment, including comparison of each edited image
with the original Image BASE"
}
**Key Reminder:**
- The "rank" array must contain exactly two values corresponding to Image 1 and Image 2.
- rank=1 indicates the better edit (higher fidelity + better execution), rank=2 indicates the
worse one.
\end{lstlisting}
\end{tcolorbox}
\begin{minipage}{\linewidth}
\captionsetup{hypcap=false}
\captionof{figure}{\textbf{Image editing evaluation system prompt.} The prompt template used to instruct the VLM judge to perform pairwise comparison for image editing, where Image BASE serves as the ground-truth reference for fidelity assessment.}
\label{fig:edit_prompt}
\end{minipage}
\clearpage

\end{document}